\documentclass[10pt,journal,compsoc]{IEEEtran}
\ifCLASSOPTIONcompsoc
\usepackage[nocompress]{cite}
\else
\usepackage{cite}
\fi

\usepackage{amsmath,amsfonts}
\usepackage{algorithmic}
\usepackage{algorithm}
\usepackage{array}
\usepackage[caption=false,font=footnotesize,labelfont=sf,textfont=sf]{subfig}
\usepackage{textcomp}
\usepackage{stfloats}
\usepackage{url}
\usepackage{verbatim}
\usepackage{graphicx}
\usepackage{cite}
\usepackage{multirow}
\usepackage{booktabs}
\usepackage{color}
\usepackage{hyperref}
\usepackage{CJKutf8}
\usepackage[figuresright]{rotating}
\usepackage[numbers,longnamesfirst]{natbib}

\begin{document}
\begin{CJK}{UTF8}{gbsn}
\title{Quantum Conflict Measurement in Decision Making for Out-of-Distribution Detection}

\author{Yilin Dong, Tianyun Zhu, Xinde Li,\IEEEmembership{~Senior Member, IEEE}, Jean Dezert, Rigui Zhou, Changming Zhu,\\ Lei Cao and Shuzhi Sam Ge, \IEEEmembership{~Fellow, IEEE}
\thanks{Yilin Dong, Tianyun Zhu, Rigui Zhou. Changming Zhu and Lei Cao are with the College of Information Engineering, Shanghai Maritime University, Shanghai 201306, China (e-mail: yldong@shmtu.edu.cn; 202330310212@stu.shmtu.edu.cn; rgzhou@shmtu.edu.cn; cmzhu@shmtu.edu.cn; lcao@shmtu.edu.cn).}
\thanks{Xinde Li is with the Key Laboratory of Measurement and Control of CSE, School of Automation, and the School of Cyber Science and Engineering, Southeast University, Nanjing 210096, China (e-mail: xindeli@seu.edu.cn).}
\thanks{Jean Dezert is with the Information Processing and Systems Department, ONERA—The French Aerospace Lab, F-91123 Palaiseau, France (e-mail: jean.dezert@onera.fr).}
\thanks{Shuzhi Sam Ge is with the Social Robotics Laboratory, Department of Electrical and Computer Engineering, Interactive Digital Media Institute, National University of Singapore, Singapore 119077, and also with the Institute for Future, Qingdao University, Qingdao 266071, China (e-mail: samge@nus.edu.sg).}
\thanks{}}
\markboth{}%
{Shell \MakeLowercase{\textit{et al.}}: A Sample Article Using IEEEtran.cls for IEEE Journals}


\IEEEtitleabstractindextext{%
\begin{abstract}
Quantum Dempster-Shafer Theory (QDST) uses quantum interference effects to derive a quantum mass function (QMF) as a fuzzy metric type from information obtained from various data sources. In addition, QDST uses quantum parallel computing to speed up computation. Nevertheless, the effective management of conflicts between multiple QMFs in QDST is a challenging question. This work aims to address this problem by proposing a Quantum Conflict Indicator (QCI) that measures the conflict between two QMFs in decision-making. Then, the properties of the QCI are carefully investigated. The obtained results validate its compliance with desirable conflict measurement properties such as non-negativity, symmetry, boundedness, extreme consistency and insensitivity to refinement. We then apply the proposed QCI in conflict fusion methods and compare its performance with several commonly used fusion approaches. This comparison demonstrates the superiority of the QCI-based conflict fusion method. Moreover, the Class Description Domain Space (C-DDS) and its optimized version, C-DDS+ by utilizing the QCI-based fusion method, are proposed to address the Out-of-Distribution (OOD) detection task. The experimental results show that the proposed approach gives better OOD performance with respect to several state-of-the-art baseline OOD detection methods. Specifically, it achieves an average increase in Area Under the Receiver Operating Characteristic Curve (AUC) of 1.2\% and a corresponding average decrease in False Positive Rate at 95\% True Negative Rate (FPR95) of 5.4\% compared to the optimal baseline method.
\end{abstract}

}

\maketitle

\section{Introduction}
\IEEEPARstart{U}{ncertainty} pervades many real-world applications, especially in approximate reasoning and decision-making processes. The analysis and management of uncertainty to facilitate decision-making in various domains has recently received considerable scientific attention. These domains include image classification \cite{jiang2020multi}, medical diagnosis \cite{cao2019extraction}, \cite{chen2024evidence}, information fusion \cite{xiao2020giq}, and decision-making \cite{herrera2020revisiting}. A plethora of theoretical frameworks has been introduced to quantify and manage uncertainty, such as fuzzy sets \cite{zadeh1965fuzzy}, rough sets \cite{pawlak1982rough}, Z-numbers \cite{zadeh2011note}, Dempster-Shafer Theory of evidence \cite{shafer1992dempster}, D-numbers \cite{deng2019total}, Entropy \cite{yager2008entropy}, \cite{khalaj2020new}, and other hybrid methodologies \cite{kang2020environmental}. Among these, Dempster-Shafer Theory of evidence (DST) is widely utilized. DST models uncertainty quantitatively through mass functions \cite{jiang2019novel}, which are informed by the appropriate degree of fuzziness associated with the uncertain variables as their quality \cite{yager1999class}, \cite{banon1981distinction}. These mass functions are classically combined using Dempster's Rule of Combination (DRC). The results generated by the DRC are quite fault-tolerant and reduce the level of uncertainty in practical applications \cite{yager2018generalized}.

However, the computational complexity of DRC grows exponentially with the dimension of the frame of discernment (FoD) \cite{deng2023novel}. Furthermore, the assumption of the reliability of the sources of evidence can yield counterintuitive results in complex scenarios \cite{deng2015generalized}. In response to these challenges, Deng proposed a controversial Quantum Dempster-Shafer Theory (QDST) \cite{pan2020a}, which is claimed to facilitate parallel computation through quantum circuits, thereby accelerating computational processes. Additionally, it models uncertainty via quantum mass functions (QMFs) and introduces angular phases to simultaneously represent propositional support information and reliability, thereby achieving diversification in propositional reliability assessment.

In light of prior investigations within DST \cite{han2016belief}, optimizing conflict resolution among multiple evidences could enhance decision-level accuracy \cite{yong2004combining}. Although numerous established methods for conflict metrics and management exist within DST \cite{jousselme2001new}, \cite{xiao2020novel}, addressing conflict metrics and management within QDST remains a main concern. Pan proposed a conflict metric method \cite{pan2020a}, while Deng extended Jousselme's distance to the QDST and introduced a conflict-based discount fusion method \cite{pan2022distance}. Both advancements have further refined the accuracy of QDST decision-making. Nevertheless, these conflict measurement methods lack certain desirable properties expected in robust conflict resolution methods, like boundedness \cite{jiang2018correlation}. Therefore, we propose to develop a conflict metric and management method that adheres to the properties detailed in \cite{jiang2018correlation} within the theoretical framework of QDST, including non-negativity, symmetry, boundedness, extreme consistency and insensitivity to refinement, thereby augmenting the scope of decision performance at the decision-making level.

Therefore, in this paper, a novel measure method of quantum conflict is proposed. We found that not only the disparity in support between identical propositions in two QMFs and the interrelationships among the elements affect their correlation, but also the variation in angular phases of the elements has an impact. Based on this insight, the main contributions of this paper are as follows:

1) Quantum Correlation Coefficient (QCC) is first proposed to measure the degree of correlation between quantum mass functions (QMFs). The QCC accounts not only for the disparity in support levels between identical propositions within the QMFs but also incorporates the interrelationships among elements and the angular phases of them into the discriminant range. Based on QCC, Quantum Conflict Indicator (QCI) is proposed to further encapsulate the relationship between correlation coefficients and conflicts, thereby measuring the level of conflicts among QMFs. The QCI exhibits desirable properties for conflict measurement and its properties is thoroughly analyzed and mathematically verified.

2) A QCI-based conflict fusion method is proposed to be applied on the UCI dataset, which exhibits increased recognition rates compared to  some of the classical methods. At the same time, in our quest to uncover the untapped potential of QCI, we explore its application in the out-of-distribution (OOD) detection task by introducing a pioneering OOD detection approach, called Class Description Domain Space (C-DDS), as well as its refined iteration, C-DDS+. Compared to numerous state-of-the-art OOD detection methods, our approach exhibits superior detection accuracy.

The remainder of this article is organized as follows: Section \uppercase\expandafter{\romannumeral2} introduces some of the preparatory knowledge required for this article. Section \uppercase\expandafter{\romannumeral3} gives a detailed description of the newly proposed QCC and QCI and their properties. Section \uppercase\expandafter{\romannumeral4} presents the QCI-based conflict fusion method and compares it with other fusion methods on the UCI dataset. Section \uppercase\expandafter{\romannumeral5} presents C-DDS and C-DDS+ and \uppercase\expandafter{\romannumeral6} compares them with other OOD detection methods on the ImageNet dataset.
\uppercase\expandafter{\romannumeral7} concludes the paper.

\section{Preliminaries}
In this section, we present a brief introduction to Quantum Dempster-Shafer Theory (QDST) that are essential for understanding the subsequent sections of this paper.

Assuming the quantum framework of discernment (QFoD) ${\left|\Theta\right\rangle=\lbrace\left|a_1\right\rangle,...,\left|a_n\right\rangle\rbrace}$ is an exclusive and non-empty set, where the elements ${\left|a_k\right\rangle,k=1,...,n}$ are mutually exclusive. A quantum mass function (QMF) $qm$ on the power set ${2^{\left|\Theta\right\rangle}=\lbrace\emptyset,A_1,...,A_{2^n-1}\rbrace}$ can be defined as ${qm\left(A_k\right)=\sqrt{\psi_k}\cdot e^{\rm{i}\theta_k}=x_k+y_k\rm{i}}$, where i denotes the imaginary unit, which is defined by its property ${\rm{i}^2=-1}$. Here, ${qm\left(\emptyset\right)=0}$ and ${\sum_{A_k\subseteq2^{\left|\Theta\right\rangle}}{|qm\left(A_k\right)|^2=1}}$ are satisfied. ${qm\left(A_k\right)}$ represents the amplitude given by Euler formula, which can be expressed as a complex number ${x_k+y_k\rm{i}}$. Specifically, ${x_k=\sqrt{\psi_k}\cdot\cos(\theta_k)}$ and ${y_k=\sqrt{\psi_k}\cdot\sin(\theta_k)}$. Moreover, ${\left|qm\left(A_k\right)\right|^2=x_k^2+y_k^2=\psi_k}$ represents the quantum probability, and $\theta_k$ represents the angular phase of $A_k$, ranging from the $0$ to $\frac{\pi}{2}$. Furthermore, Dempster Rule of Combination of QMFs (DRC-QM) is defined as:
\begin{equation}
	qm\left(A_k\right) = \frac{qm'\left(A_k\right)}{\sqrt{1-K}},\label{DRC-QM}
\end{equation}

\begin{equation}
	qm'\left(A_k\right)=\left\{
	\begin{aligned}
		\sum_{A_i\cap A_j=A_k}{qm_1\left(A_i\right)\times qm_2\left(A_j\right)},A_k\neq\emptyset\\
		0,A_k=\emptyset
	\end{aligned}
	\right.
\end{equation}

where the conflict coefficient $K$ is defined as:

\begin{equation}
	K=1-\sum_{A_i\cap A_j=\emptyset}{|qm_1\left(A_i\right)\times qm_2\left(A_j\right)|}.
\end{equation}
Note that if ${\theta_k=0}$, QDST \cite{pan2020a} degenerates to DST.

\section{A novel Quantum Correlation Coefficient and Quantum Conflict Indicator}
In this section, we introduce the Quantum Conflict Indicator (QCI) as a means to achieve a robust and effective evaluation of conflict in quantum frameworks. We first introduce the Quantum Correlation Coefficient (QCC) to measure the degree of correlation between QMFs. Building on the QCC, we then introduce the QCI as a quantitative measure for assessing conflicts between QMFs.\\
\subsection{QCC between QMFs}
\textit{Definition 1:} There are two QMFs $qm_1$ and $qm_2$ on $2^{\left|\Theta\right\rangle}$, where $2^{\left|\Theta\right\rangle}=\lbrace\emptyset,A_1,...,A_{2^n-1}\rbrace$ is the power set of QFoD $\left|\Theta\right\rangle$ and $A_k$ is non-empty and exclusive element belonging to $2^{\left|\Theta\right\rangle}$. The Quantum Correlation Coefficient (QCC) is defined as:
\begin{eqnarray}
	QCC(qm_1,qm_2)=\left[\frac{|\left\langle qm_1\middle|qm_2\right\rangle|}{||qm_1||\cdot||qm_2||}\right]^{4},\label{QCC}
\end{eqnarray}
where $\left\langle qm_1\middle|qm_2\right\rangle$ represents the inner product of $qm_1$ and $qm_2$, which is defined as:
\begin{equation}
	\left\langle qm_1\middle|qm_2\right\rangle=\sum_{i=1}^{2^n-1}\sum_{j=1}^{2^n-1}{\overline{qm_1(A_i)}qm_2(A_j)\mathcal{G}(\dot{\theta}_{i,j})\frac{|A_i\cap A_j|}{|A_i\cup A_j|}},\label{innerproduct}
\end{equation}
\begin{eqnarray}
	\mathcal{G}(\dot{\theta}_{i,j})=\frac{cos(\dot{\theta}_{i,j})+1}{2}.
\end{eqnarray}
Here, $\overline{qm_1(A_i)}$ represents the conjugate of $qm_1(A_i)$, $\dot{\theta}_{i,j}$ denotes the difference in angular phases between $A_i$ and $A_j$, and $|\cdot|$ represents the cardinality of a set. Furthermore, $||qm_1||$ represents the paradigm of $qm_1$, which is defined as:
\begin{align}
	&||qm_1||=\left[|\left\langle qm_1\middle|qm_1\right\rangle|\right]^\frac{1}{2}\notag\\
	&=\left[|\sum_{i=1}^{2^n}\sum_{j=1}^{2^n}{\overline{qm_1(A_i)}qm_1(A_j)\mathcal{G}(A_i,A_j)\frac{|A_i\cap A_j|}{|A_i\cup A_j|}}|\right]^\frac{1}{2}.
\end{align}\\
QCC has the following properties:
\begin{enumerate}
	\item{\textit{Non-negativity:} $QCC(qm_1,qm_2)$ is not negative.}
	\item{\textit{Nondegeneracy:} $QCC(qm_1,qm_2)=1$, iff $qm_1=qm_2$.}
	\item{\textit{Symmetry:} $QCC(qm_1,qm_2)=QCC(qm_2,qm_1)$.}
	\item{\textit{Boundedness:} $0\leq QCC(qm_1,qm_2)\leq 1$.}
\end{enumerate}

The proofs of the properties of QCC can be found in the Appendix \ref{QCC_p}. Note that a larger value of $QCC(qm_1,qm_2)$ indicates a higher correlation between the QMFs. Therefore, if ${QCC(qm_1,qm_2)=1}$, it implies a complete correlation between $qm_1$ and $qm_2$; whereas if ${QCC(qm_1,qm_2)=0}$, $qm_1$ and $qm_2$ are completely uncorrelated. Moreover, when quantum mass function degenerates to the classic mass function, QCC degenerates to the ECC proposed by Xiao in \cite{xiao2020novel}. For a better understanding of the properties of QCC, an example illustration is provided here.\\

\vspace{0.2em}\noindent\textbf{Example 1} Assuming that there are two QMFs, $qm_1$ and $qm_2$, on $2^{\left|\Theta\right\rangle}$:
\begin{align}
	&qm_1: qm_1(\left|a_1\right\rangle)=\sqrt{\psi}, qm_1(\vartheta)=\sqrt{(1-\psi)}e^{\rm{i}\theta_{\vartheta}}\notag\\
	&qm_2: qm_2(\left|a_1\right\rangle)=\sqrt{(1-\psi)}, qm_2(\vartheta)=\sqrt{\psi}\notag
\end{align}

\begin{figure*}[tp]
	\centering
	\subfloat[]{\includegraphics[width=2in]{E1_1.pdf}%
		\label{fig11}}
	\hfil
	\subfloat[$\theta_{\vartheta}=0$]{\includegraphics[width=1.5in]{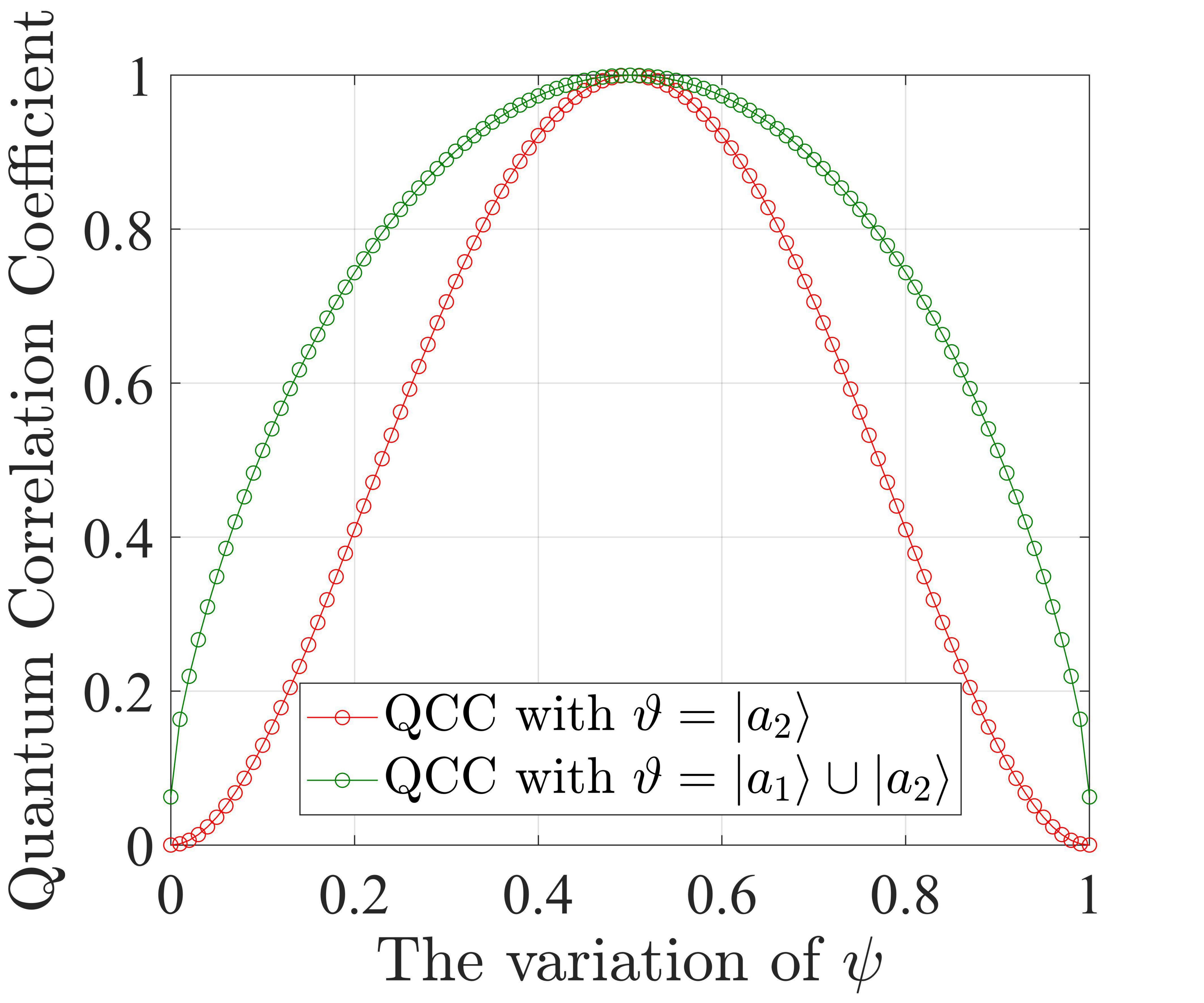}%
		\label{fig12}}
	\hfil
	\subfloat[$\psi=0.5$]{\includegraphics[width=1.5in]{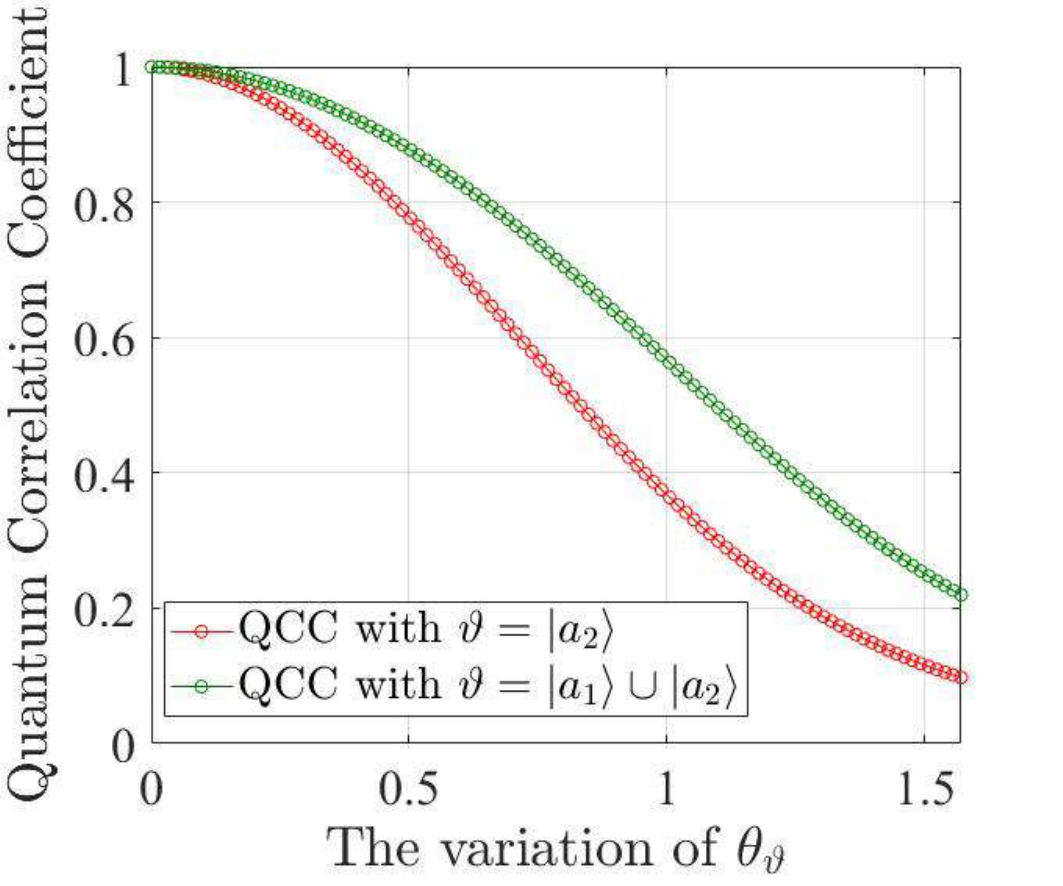}%
		\label{fig13}}
	\caption{The Quantum Correlation Coefficient in Example 1.}
	\label{Example1_QCC}
\end{figure*}

In this example, $\vartheta$ means ${\left|a_2\right\rangle}$ or ${\left|a_1\right\rangle\cup\left|a_2\right\rangle}$. $qm_1$ and $qm_2$ vary based on the value of $\psi$, $\theta_{\vartheta}$ and $\vartheta$. Here $\psi$ is constrained within the interval $[0,1]$, $\theta_{\vartheta}$ is limited to $[0,\frac{\pi}{2}]$. The corresponding QCC are shown in Fig. \ref{Example1_QCC}.

When ${\theta_{\vartheta}=0}$ and ${\psi=0.5}$, it can be observed that ${qm_1(\left|a_1\right\rangle)=qm_2(\left|a_1\right\rangle)=qm_1(\vartheta)=qm_2(\vartheta)=\frac{1}{\sqrt{2}}}$. Regardless of whether the subset of $\vartheta$ is $\left|a_2\right\rangle$ or ${\left|a_1\right\rangle\cup\left|a_2\right\rangle}$, the $QCC(qm_1,qm_2)$ attains its maximum value of $1$ because $qm_1$ and $qm_2$ are identical, indicating complete correlation between $qm_1$ and $qm_2$. 

Furthermore, for ${\psi=0}$ and $\vartheta$ is $\left|a_2\right\rangle$, it can be observed that ${qm_1(\left|a_1\right\rangle)=qm_2(\vartheta)=0}$, ${qm_1(\vartheta)=e^{\rm{i}\theta_{\vartheta}}}$ and ${qm_2(\left|a_1\right\rangle)=1}$. Conversely, for ${\psi=1}$ and $\vartheta$ is ${\left|a_1\right\rangle\cup\left|a_2\right\rangle}$, it can be observed that ${qm_1(\left|a_1\right\rangle)=qm_2(\vartheta)=1}$ and ${qm_1(\vartheta)=qm_2(\left|a_1\right\rangle)=0}$. Under these two cases, the $QCC(qm_1,qm_2)$ attains its minimum value of $0$ since $qm_1$ and $qm_2$ are different, indicating complete lack of correlation between them. Therefore, the boundedness and non-negativity of the QCC are verified.

Moreover, when ${\theta_{\vartheta}=0}$, ${\psi=0}$ and $\vartheta$ is $\left|a_2\right\rangle$, we find that ${qm_1(\left|a_1\right\rangle)=qm_2(\vartheta)=0}$ and ${qm_1(\vartheta)=qm_2(\left|a_1\right\rangle)=1}$. Similarly, when ${\theta_{\vartheta}=0}$, ${\psi=1}$ and $\vartheta$ is ${\left|a_1\right\rangle\cup\left|a_2\right\rangle}$, we have ${qm_1(\left|a_1\right\rangle)=qm_2(\vartheta)=1}$ and ${qm_1(\vartheta)=qm_2(\left|a_1\right\rangle)=0}$. Under these two cases, the ${QCC(qm_1,qm_2)=0.25}$. This result is reasonable because there is an intersection element $\left|a_2\right\rangle$ between $\left|a_2\right\rangle$ and ${\left|a_1\right\rangle\cup\left|a_2\right\rangle}$. Consequently, these two QMFs are not completely uncorrelated, which justifies the value of $QCC(qm_1,qm_2)$ being $0.25$ instead of $0$.

Furthermore, as $\psi$ increases from $0$ to $0.5$, the value of $QCC(qm_1,qm_2)$ progressively increases for any $\theta_{\vartheta}$ both when $\vartheta$ is $\left|a_2\right\rangle$ and $\vartheta$ is ${\left|a_1\right\rangle\cup\left|a_2\right\rangle}$. This outcome aligns with the expected behavior since $qm_1$ and $qm_2$ become more similar as $\psi$ increases from $0$ to $0.5$. Conversely, as $\psi$ increases from $0.5$ to $1$, the value of $QCC(qm_1,qm_2)$ gradually decreases for any $\theta_{\vartheta}$ both when $\vartheta$ is $\left|a_2\right\rangle$ and $\vartheta$ is ${\left|a_1\right\rangle\cup\left|a_2\right\rangle}$. Again, This aligns with the anticipated result as $qm_1$ and $qm_2$ become dissimilar when $\psi$ increases from $0.5$ to $1$. Note that Fig. \ref{Example1_QCC}(a) exhibits symmetry about ${\psi = 0.5}$, thereby confirming the symmetry of the QCC.

Additionally, for ${\psi=0.5}$, it can be observed that ${qm_1(\left|a_1\right\rangle)=\frac{1}{\sqrt{2}}}$, ${qm_1(\vartheta)=\frac{1}{\sqrt{2}}e^{\rm{i}\theta_{\vartheta}}}$, ${qm_2(\left|a_1\right\rangle)=\frac{1}{\sqrt{2}}}$ and ${qm_2(\vartheta)=\frac{1}{\sqrt{2}}}$. In this case, as $\theta_{\vartheta}$ increases from 0.5 to 1, $QCC(qm_1,qm_2)$ gradually decreases both when $\vartheta$ is $\left|a_2\right\rangle$ and $\vartheta$ is ${\left|a_2\right\rangle\cup\left|a_2\right\rangle}$. This result can be attributed to the fact that as $\theta_{\vartheta}$ increases, $qm_1(\vartheta)$ and $qm_2(\vartheta)$ become dissimilar, resulting in a smaller correlation coefficient between $qm_1$ and $qm_2$. In particular, due to the existence of the intersection element $\left|a_2\right\rangle$ between $\left|a_2\right\rangle$ and ${\left|a_1\right\rangle\cup\left|a_2\right\rangle}$, $QCC(qm_1,qm_2)$ with ${\vartheta=\left|a_2\right\rangle}$ can never exceed $QCC(qm_1,qm_2)$ with ${\vartheta=\left|a_1\right\rangle\cup\left|a_2\right\rangle}$.

\subsection{QCI between QMFs}
\textit{Definition 8:} For two QMFs $qm_1$ and $qm_2$ on $2^{\left|\Theta\right\rangle}$, the Quantum Conflict Indicator (QCI) is defined as:
\begin{eqnarray}
	QCI(qm_1,qm_2)=1-QCC(qm_1,qm_2).\label{QCI}
\end{eqnarray}

The QCI has desirable properties required for a conflict measurement \cite{jiang2018correlation}, including:
\begin{enumerate}
	\item{\textit{Non-negativity:} $QCI(qm_1,qm_2)$ is not negative.}
	\item{\textit{Symmetry:} $QCI(qm_1,qm_2)=QCI(qm_2,qm_1)$.}
	\item{\textit{Boundedness:} $0\leq QCI(qm_1,qm_2)\leq 1$.}
	\item{\textit{Extreme consistency:} 1) $QCI(qm_1,qm_2)=1$, iff for $A_i$ and $ A_j$ of $qm_1$ and $qm_2$ respectively, $(\cup A_i)\cap(\cup A_j)=\emptyset$; 2) $QCI(qm_1,qm_2)=0$, iff $qm_1=qm_2$.}
	\item{\textit{Insensitivity to refinement:} for $qm_1$ and $qm_2$ refined from $2^{\left|\Theta\right\rangle_1}$ to $2^{\left|\Theta\right\rangle_2}$, $QCI(qm_1,qm_2)$ in $2^{\left|\Theta\right\rangle_1}$ is equal to $QCI(qm_2,qm_1)$ in $2^{\left|\Theta\right\rangle_2}$.}
\end{enumerate}

The proofs of the properties of QCI are given in the Appendix \ref{QCI_p}. Note that a larger value of $QCI(qm_1,qm_2)$ indicates a higher conflict between the QMFs. Therefore, if ${QCI(qm_1,qm_2)=1}$, it implies a complete conflict between $qm_1$ and $qm_2$, which means two QMFs support completely different propositions; whereas if ${QCI(qm_1,qm_2)=0}$, $qm_1$ and $qm_2$ are completely non-conflicting, which means two QMFs support same propositions. To better understand the properties of QCI, some examples are provided here.\\

\vspace{0.2em}\noindent\textbf{Example 2} Assuming that there are two QMFs, $qm_1$ and $qm_2$, on $2^{\left|\Theta\right\rangle}$:
\begin{align}
	&qm_1: qm_1(\left|a_1\right\rangle)=\frac{\ \ x+y\rm{i}\ \ }{d(x,y)},qm_1(\vartheta_1)=\frac{1-x+y\rm{i}}{d(x,y)}\notag\\
	&qm_2: qm_2(\left|a_1\right\rangle)=\frac{1-x+y\rm{i}}{d(x,y)},qm_2(\vartheta_2)=\frac{\ \ x+y\rm{i}\ \ }{d(x,y)}\notag
\end{align}

Example 2 is obtained from \cite{pan2022distance}. Note that in this paper, the angular phase is constrained within the interval $[0,\frac{\pi}{2}]$, and both the real and imaginary parts are positive. In this example, $d(x,y)$ is $\sqrt{x^2+2y^2+(1-x)^2}$. $qm_1$ and $qm_2$ vary based on the value of $x$, $y$ and the subset of $\vartheta_1$, $\vartheta_2$. Here $x$ and $y$ are limited to the interval $[0,1]$.
\begin{figure}[!h]
	\centering
	\includegraphics[width=2.2in]{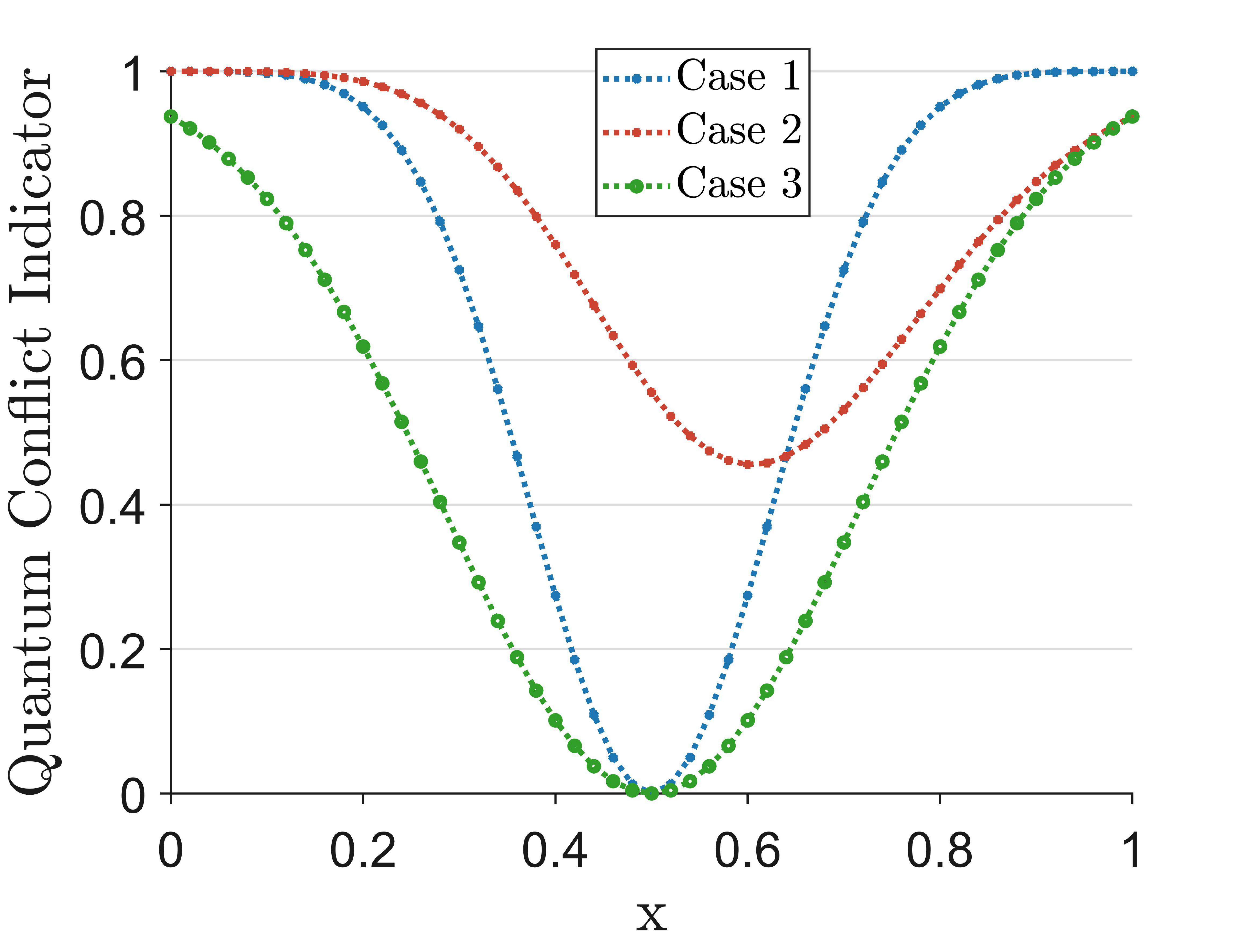}%
	\caption{The Quantum Conflict Indicator in Example 2 when $y=0$.}
	\label{E2}
\end{figure}

\begin{figure*}[tp]
	\centering
	\subfloat[$\vartheta_1=\vartheta_2=\left|a_2\right\rangle$]{\includegraphics[width=1.8in]{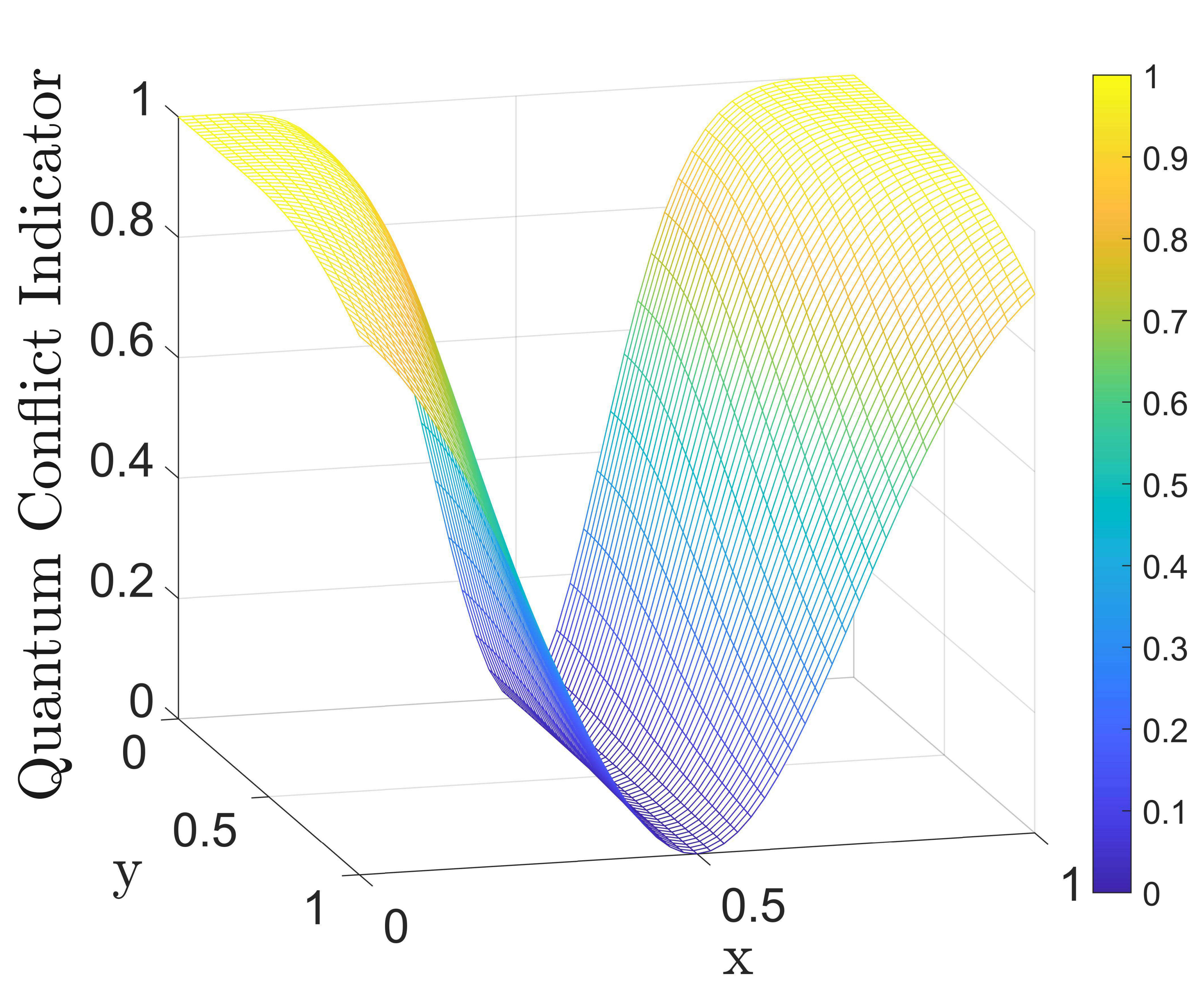}%
		\label{fig21}}
	\hfil
	\subfloat[$\vartheta_1=\left|a_2\right\rangle$, $\vartheta_2=\left|a_1\right\rangle\cup\left|a_2\right\rangle$]{\includegraphics[width=1.8in]{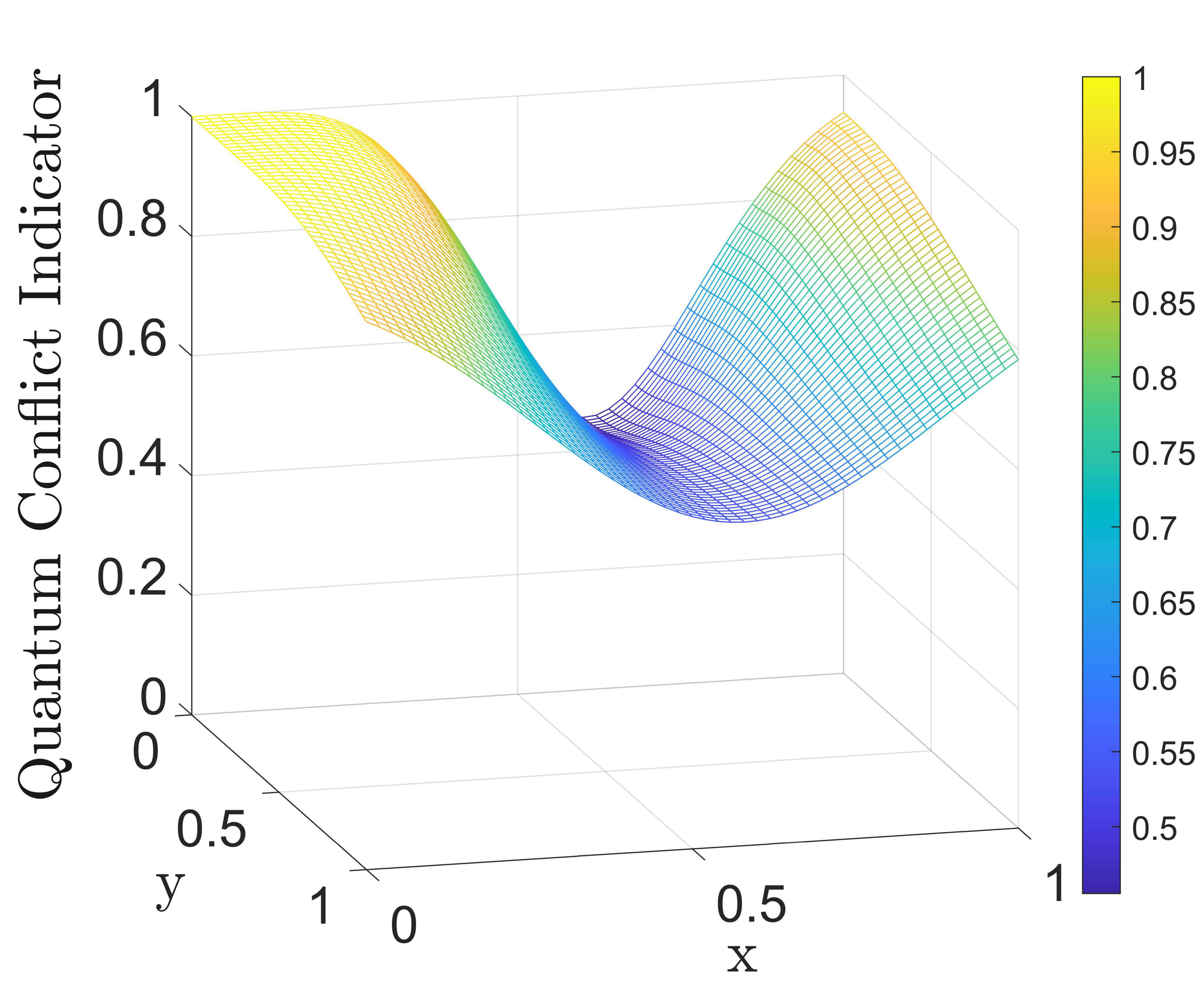}%
		\label{fig22}}
	\hfil
	\subfloat[$\vartheta_1=\vartheta_2=\left|a_1\right\rangle\cup\left|a_2\right\rangle$]{\includegraphics[width=1.8in]{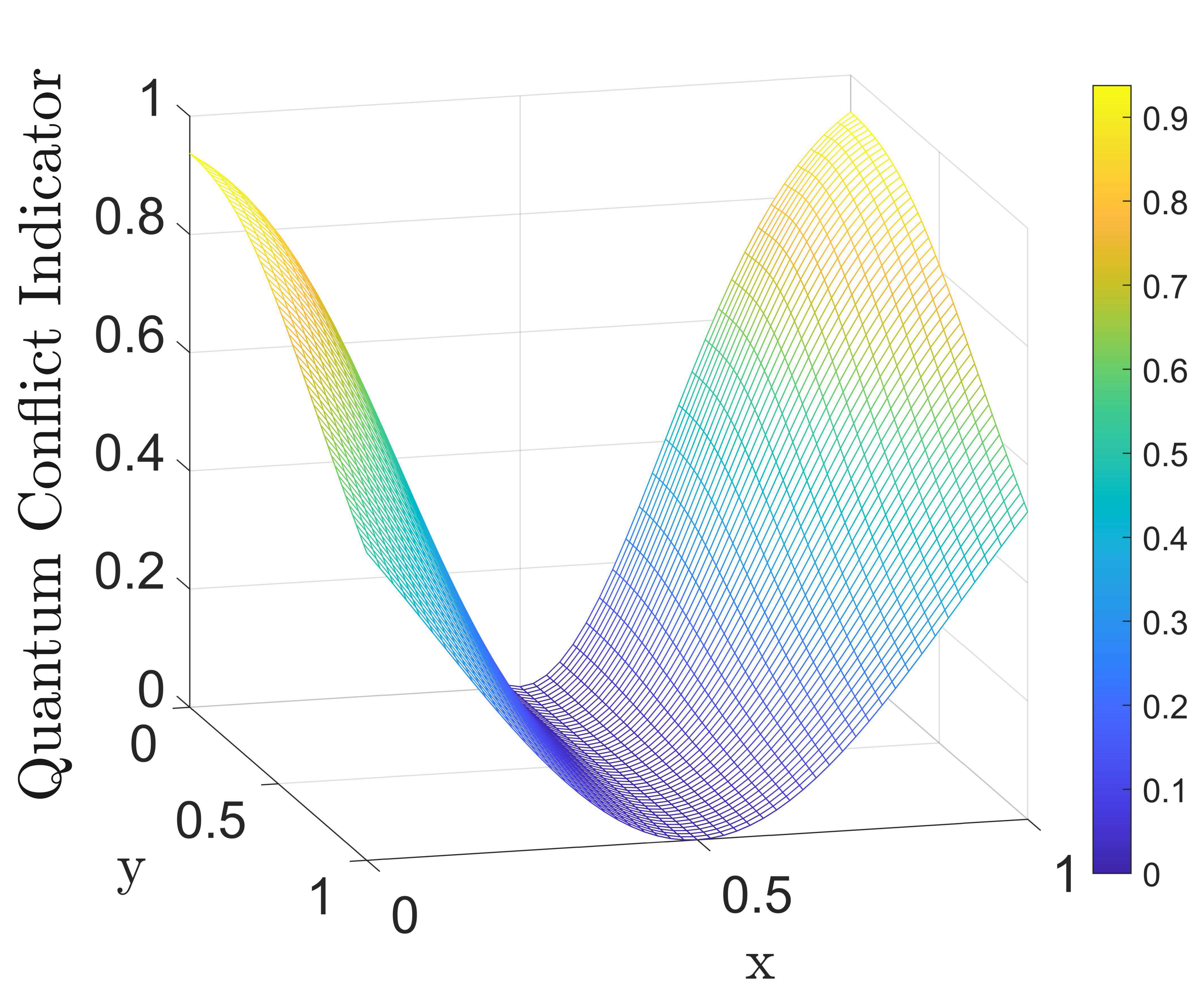}%
		\label{fig23}}
	\caption{The Quantum Conflict Indicator in Example 2.}
	\label{Example2}
\end{figure*}

\begin{figure*}[htbp]
	\centering
	\begin{minipage}{0.235\linewidth}
		\centering
		\subfloat[$\psi_2=0.3$]{\includegraphics[width=1.53in]{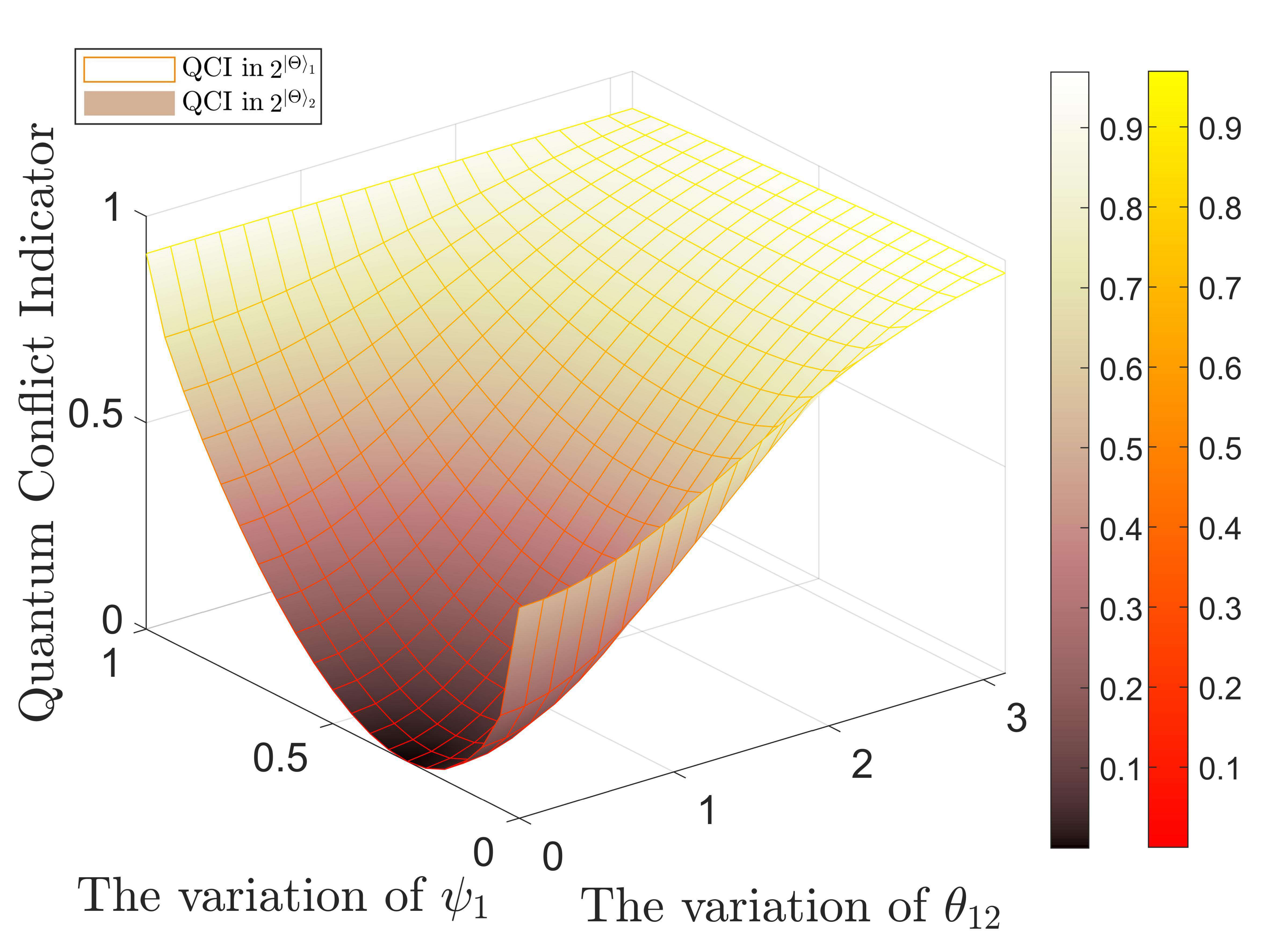}%
			\label{fig31}}
	\end{minipage}
	\begin{minipage}{0.235\linewidth}
		\centering
		\subfloat[$\psi_2=0.5$]{\includegraphics[width=1.53in]{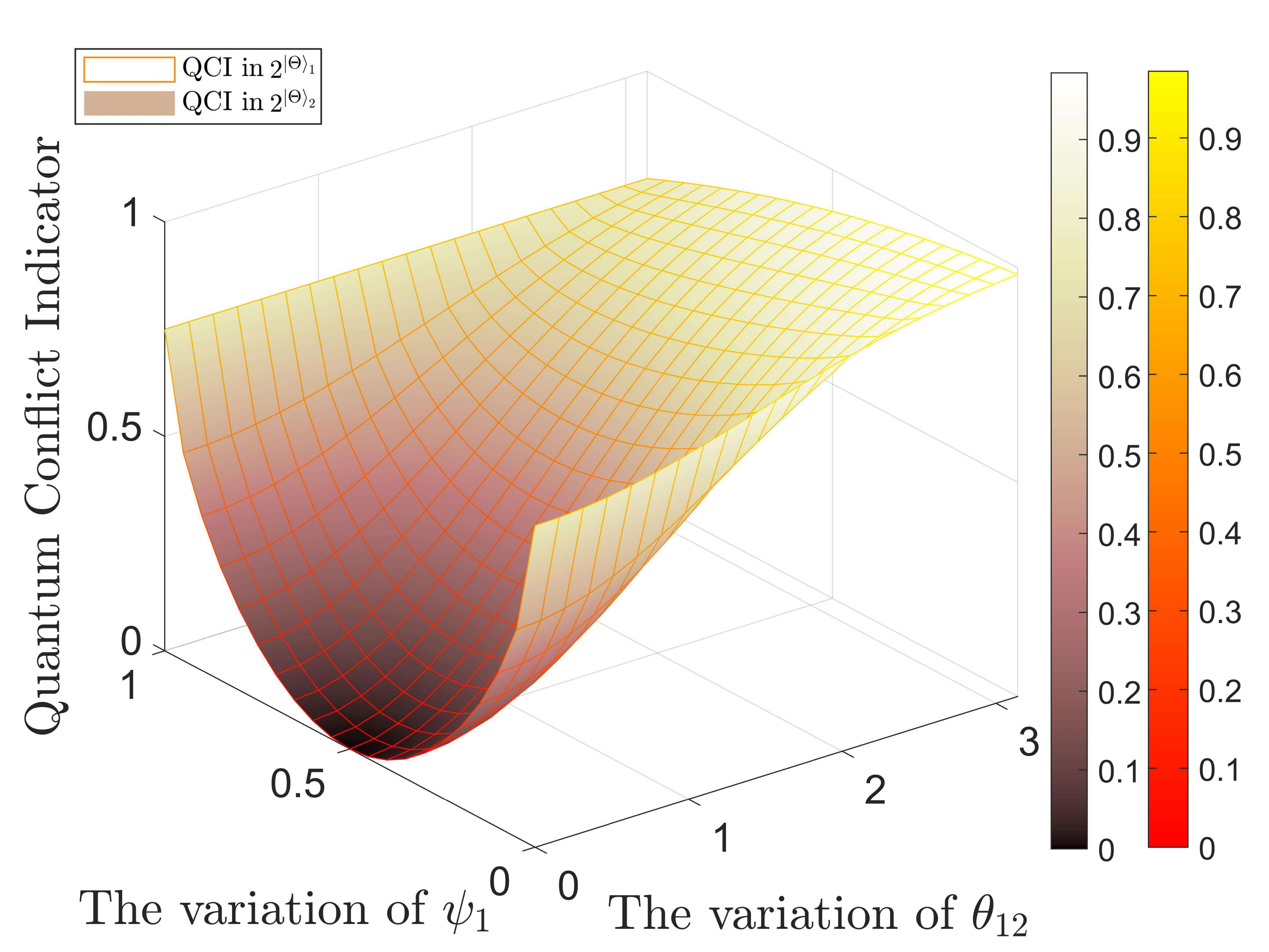}%
			\label{fig32}}
	\end{minipage}
	\begin{minipage}{0.235\linewidth}
		\centering
		\subfloat[$\psi_2=0.7$]{\includegraphics[width=1.5in]{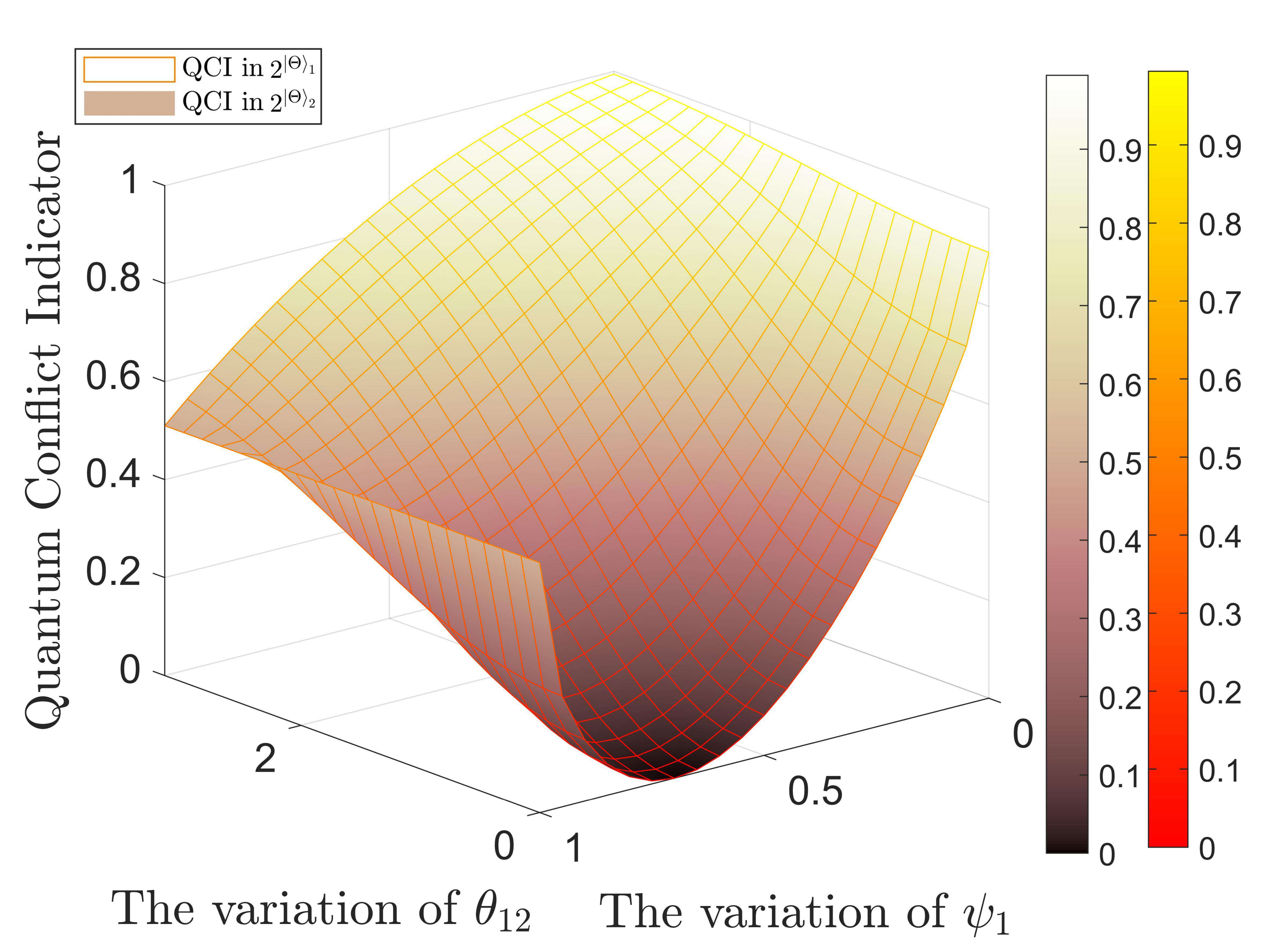}%
			\label{fig33}}
	\end{minipage}
	\begin{minipage}{0.235\linewidth}
		\centering
		\subfloat[$\psi_2=1$]{\includegraphics[width=1.5in]{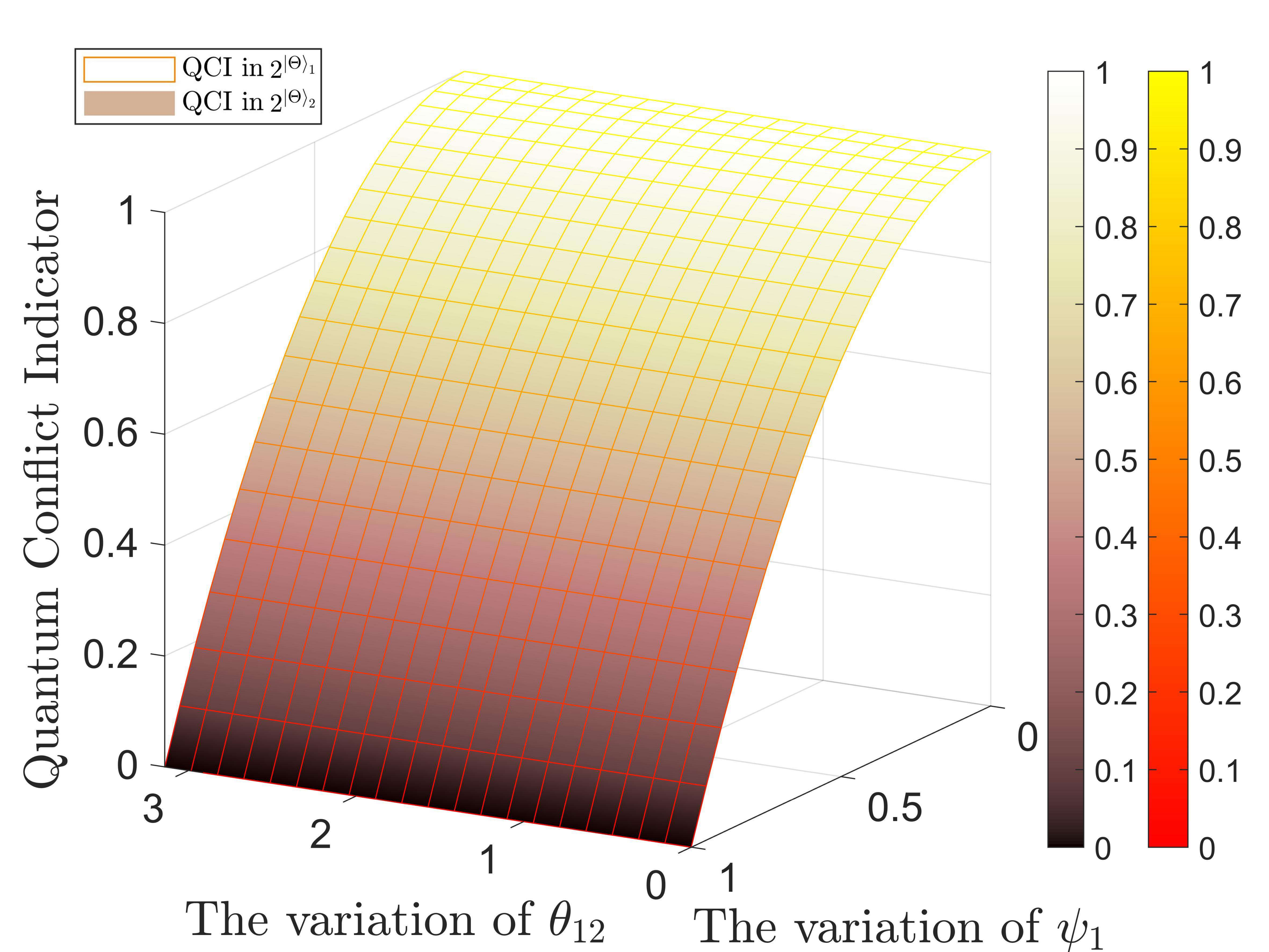}%
			\label{fig34}}
	\end{minipage}
	\qquad
	\begin{minipage}{0.235\linewidth}
		\centering
		\subfloat[$\psi_2=0.3$]{\includegraphics[width=1.5in]{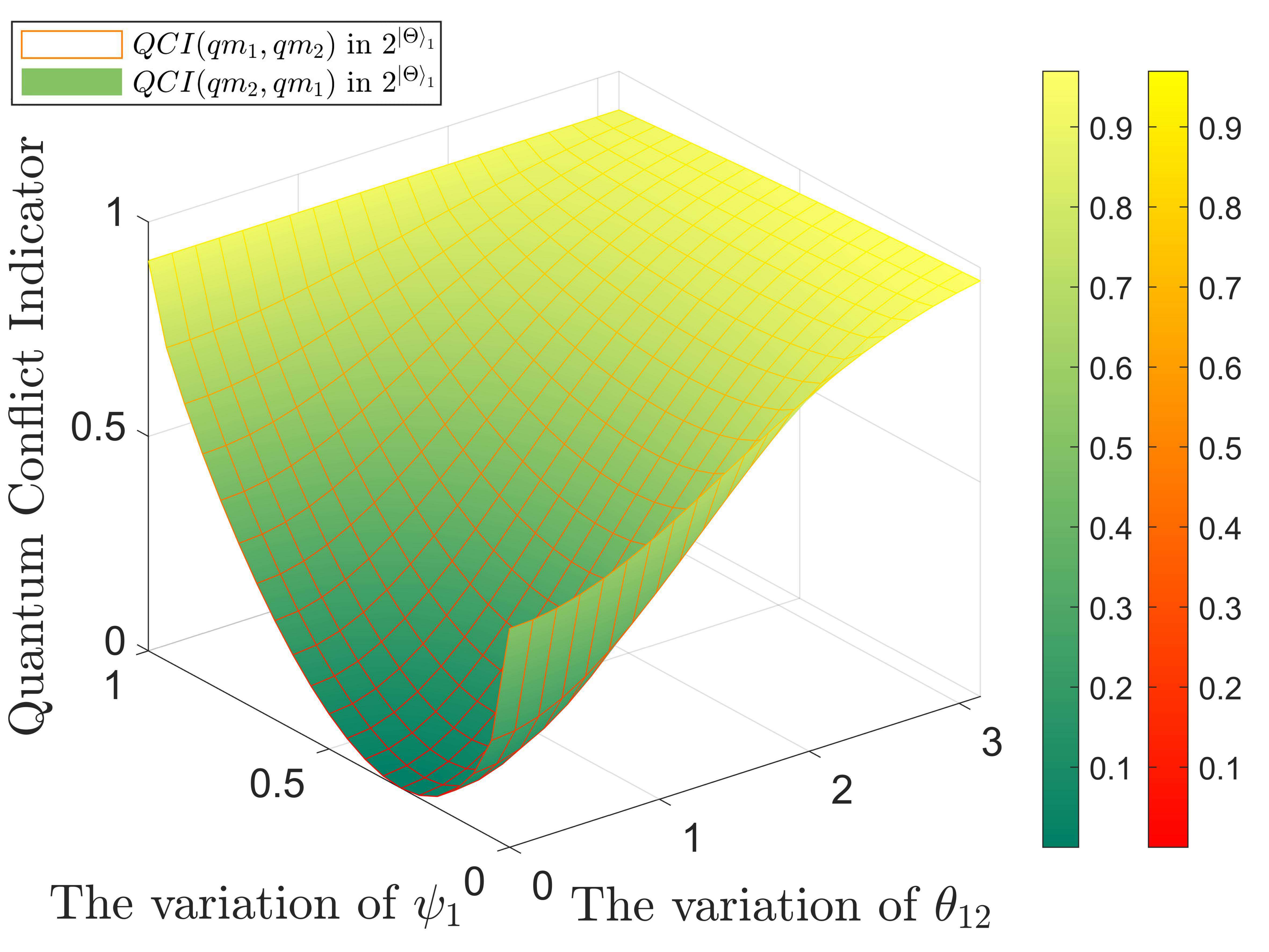}%
			\label{fig35}}
	\end{minipage}
	\begin{minipage}{0.235\linewidth}
		\centering
		\subfloat[$\psi_2=0.5$]{\includegraphics[width=1.5in]{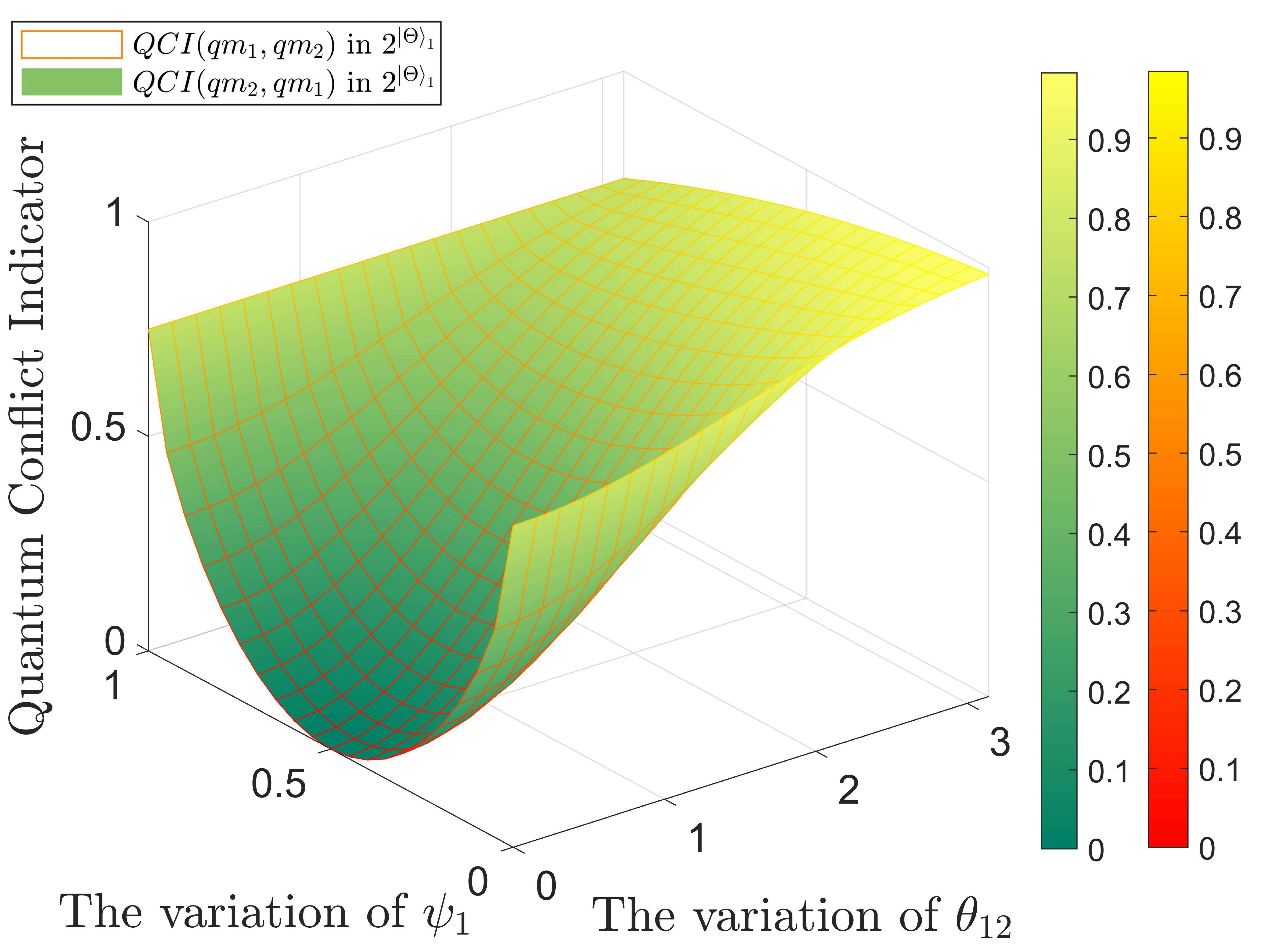}%
			\label{fig36}}
	\end{minipage}
	\begin{minipage}{0.235\linewidth}
		\centering
		\subfloat[$\psi_2=0.7$]{\includegraphics[width=1.5in]{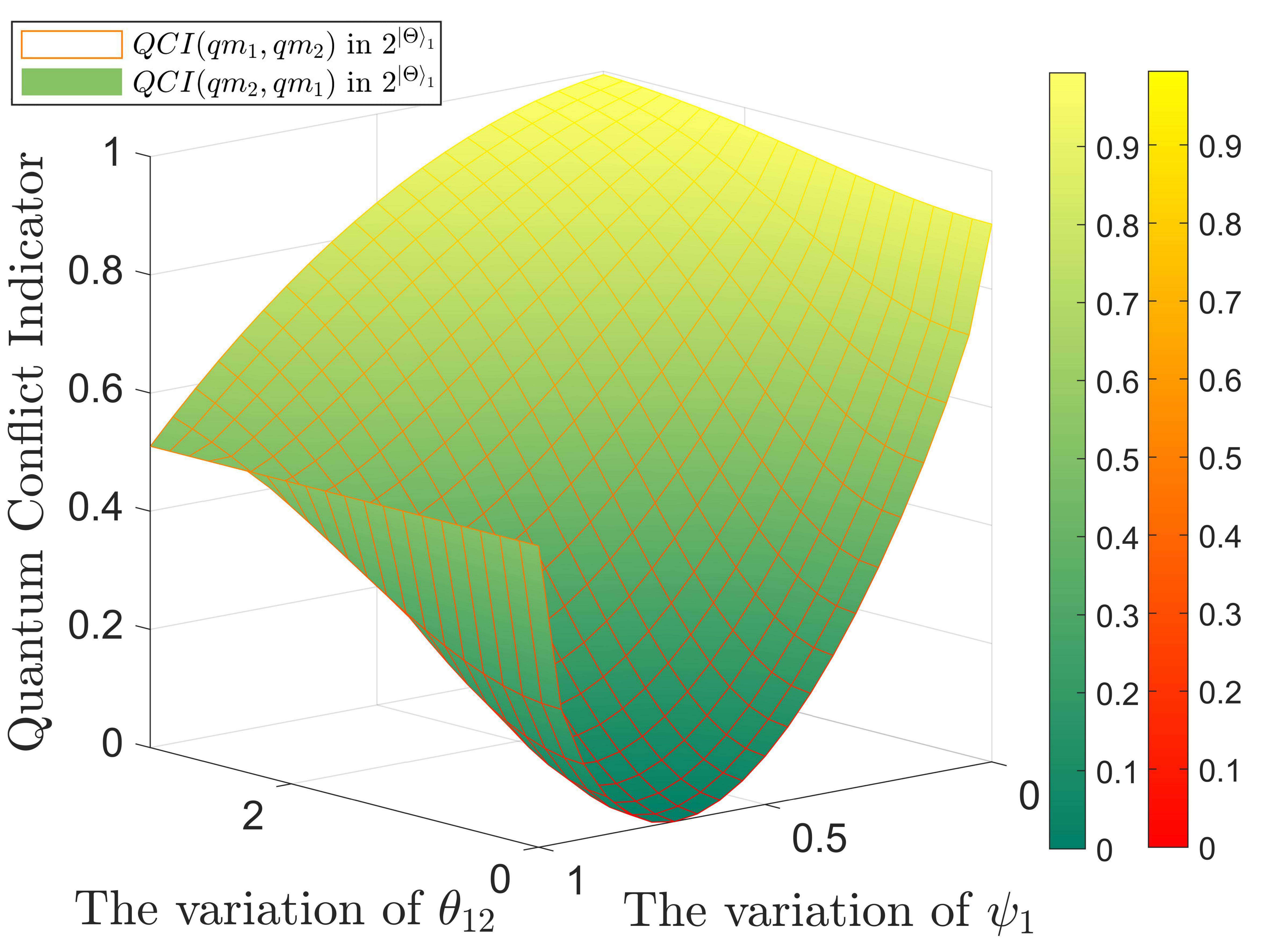}%
			\label{fig37}}
	\end{minipage}
	\begin{minipage}{0.235\linewidth}
		\centering
		\subfloat[$\psi_2=1$]{\includegraphics[width=1.45in]{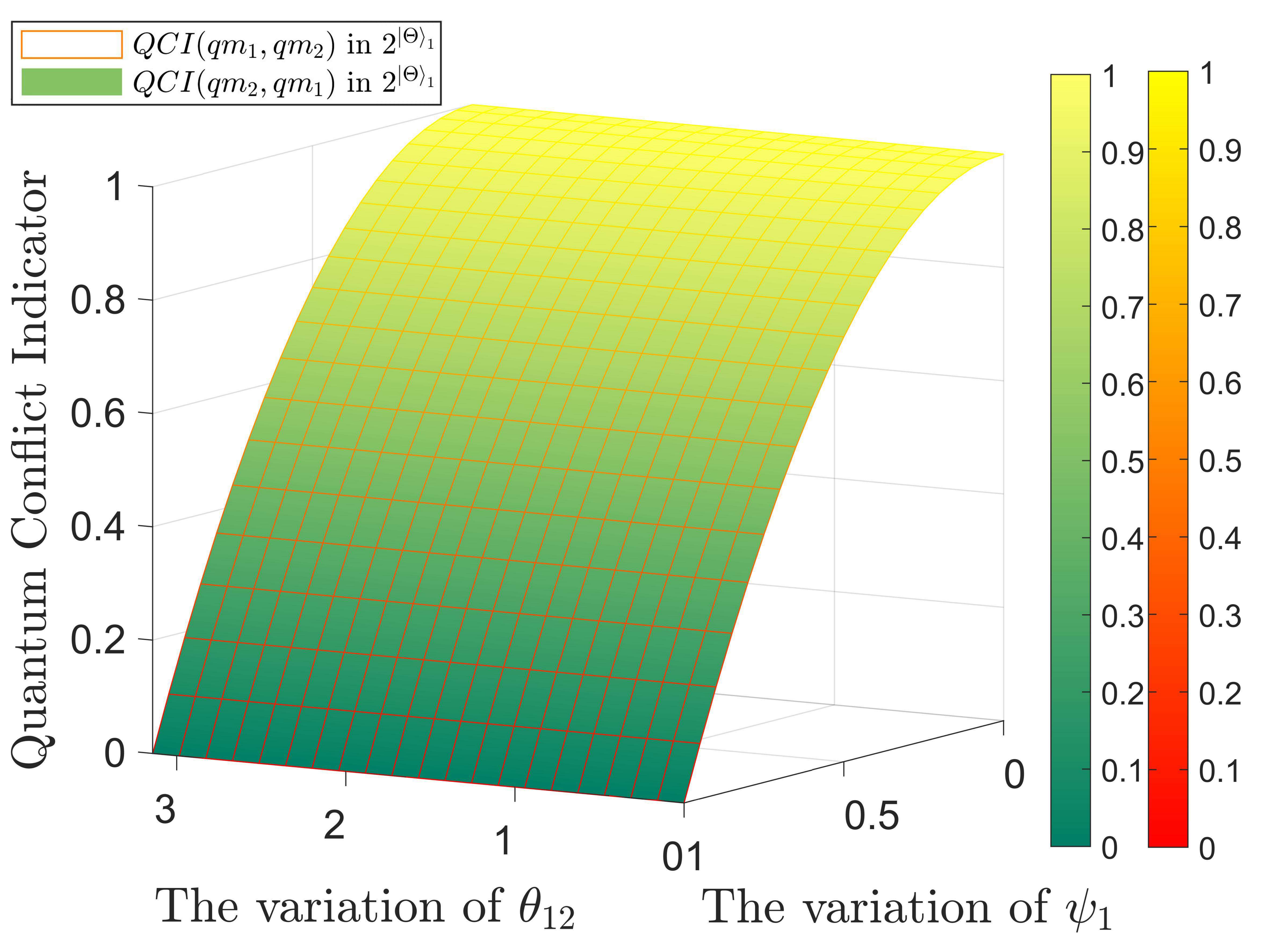}%
			\label{fig38}}
	\end{minipage}
	\caption{The Quantum Conflict Indicator in Example 3.}
	\label{Example3}
\end{figure*}

When ${y=0}$, the quantum mass function degenerates to the classic mass function and the QCI between $qm_1$ and $qm_2$ is shown in Fig. \ref{E2}. Notably, regardless of how $x$ varies, the value of QCI is never less than $0$, thus satisfying the non-negativity. Three cases are discussed based on variations in $\vartheta_1$ and $\vartheta_2$:

\textbf{Case 1 (${\vartheta_1=\vartheta_2=\left|a_2\right\rangle}$):} when ${x=0.5}$, we observe that $qm_1(\left|a_1\right\rangle)=qm_1(\vartheta_2)=qm_2(\left|a_1\right\rangle)=qm_2(\vartheta_2)=\frac{1}{\sqrt{2}}$. In this case, the  $QCI(qm_1,qm_2)$ attains its minimum value of $0$ because $qm_1$ and $qm_2$ are complete non-conflicting. Moreover, for ${x=0}$, we have ${qm_1(\left|a_1\right\rangle)=qm_2(\vartheta_2)=0}$ and ${qm_1(\vartheta_1)=qm_2(\left|a_1\right\rangle)=1}$; while for ${x=1}$, we have ${qm_1(\vartheta_1)=qm_2(\left|a_1\right\rangle)=0}$ and ${qm_1(\left|a_1\right\rangle)=qm_2(\vartheta_2)=1}$. Under these two cases, the $QCI(qm_1,qm_2)$ attains its maximum value of $1$ because $qm_1$ and $qm_2$ are complete conflicting. 

\textbf{Case 2 (${\vartheta_1=\left|a_2\right\rangle}$, ${\vartheta_2=\left|a_1\right\rangle\cup\left|a_2\right\rangle}$):} when ${x=0}$, we have ${qm_1(\left|a_1\right\rangle)=qm_2(\vartheta_2)=0}$ and ${qm_1(\vartheta_1)=qm_2(\left|a_1\right\rangle)=1}$. In this case, $qm_1$ and $qm_2$ are completely conflicting, resulting in ${QCI(qm_1,qm_2)=1}$. Furthermore, when ${x=1}$, we have ${qm_1(\vartheta_1)=qm_2(\left|a_1\right\rangle)=0}$ and ${qm_1(\left|a_1\right\rangle)=qm_2(\vartheta_2)=1}$. In this case, the ${QCI(qm_1,qm_2)=0.9375}$, indicating that $qm_1$ and $qm_2$ are not completely conflicting due to the existence of the intersection element $\left|a_1\right\rangle$ between $\vartheta_1$ and $\vartheta_2$.

\textbf{Case 3 (${\vartheta_1=\vartheta_2=\left|a_1\right\rangle\cup\left|a_2\right\rangle}$):} when ${x=0}$, we have ${qm_1(\left|a_1\right\rangle)=qm_2(\vartheta_2)=0}$ and ${qm_1(\vartheta_1)=qm_2(\left|a_1\right\rangle)=1}$; while for ${x=1}$, we have ${qm_1(\vartheta_1)=qm_2(\left|a_1\right\rangle)=0}$ and ${qm_1(\left|a_1\right\rangle)=qm_2(\vartheta_2)=1}$. Additionally, when ${x=0.5}$, the $QCI(qm_1,qm_2)$ attains its minimum value of $0$. Under these two cases, the values of $QCI(qm_1,qm_2)$ are all $0.9375$ due to the existence of the intersection element $\left|a_1\right\rangle$ between $\vartheta_1$ and $\vartheta_2$. Notably, the $QCI(qm_1,qm_2)$ in Case 3 is always less than the $QCI(qm_1,qm_2)$ in Case 1 due to the existence of the intersection element.

After considering the value of $x$ and $y$, the QCI between $qm_1$ and $qm_2$ is shown in Fig. \ref{Example2}. Three cases are discussed based on variations in $\vartheta_1$ and $\vartheta_2$:

\textbf{Case 1 (${\vartheta_1=\vartheta_2=\left|a_2\right\rangle}$):} in Fig. \ref{Example2}(a), the symmetry of the QCI with respect to changes in $x$ and $y$ is evident. When ${x=0.5}$, we have $qm_1(\left|a_1\right\rangle)=qm_2(\left|a_1\right\rangle)=qm_1(\vartheta_1)=qm_2(\vartheta_2)=\frac{0.5}{d(0.5,y)}+\frac{y\rm{i}}{d(0.5,y)}$. In this case, the $QCI(qm_1,qm_2)$ attains its minimum value of $0$ in all $y$. Likewise, when ${x=0}$, ${y=0}$ or ${x=1}$, ${y=0}$, in this case, the quantum mass function degenerates to the classic mass function and the $QCI(qm_1,qm_2)$ attains its maximum value of $1$. Obviously, the QCI satisfies the boundedness and extreme consistency. Moreover, when $x$ is fixed, the $QCI(qm_1,qm_2)$ decreases with increasing values of $y$, indicating the important role played by the imaginary number component in the QCI.

\textbf{Case 2 (${\vartheta_1=\left|a_2\right\rangle}$, ${\vartheta_2=\left|a_1\right\rangle\cup\left|a_2\right\rangle}$):} in Fig. \ref{Example2}(b), the variation of QCI with $x$ and $y$ differs Case 1 due to the existence of the intersection element $\left|a_1\right\rangle$ between $\vartheta_1$ and $\vartheta_2$. Furthermore, the QCI is always greater than $0$ regardless of changes in $x$ and $y$.

\textbf{Case 3 (${\vartheta_1=\vartheta_2=\left|a_1\right\rangle\cup\left|a_2\right\rangle}$):} in Fig. \ref{Example2}(b), the variation of QCI with $x$ and $y$ is similar to Case 1. However, due to the existence of the intersection element $\left|a_1\right\rangle$ between $\vartheta_1$ and $\vartheta_2$, The relationship between $qm_1$ and $qm_2$ transitions from complete conflict in Case 1 to partial conflict in Case 3. This leads to the result that the $QCI(qm_1,qm_2)$ in Case 3 is never greater than the $QCI(qm_1,qm_2)$ in Case 1.\\

\vspace{0.2em}\noindent\textbf{Example 3} Assuming that there are two QMFs, $qm_1$ and $qm_2$, on $2^{\left|\Theta\right\rangle_1}$ and $2^{\left|\Theta\right\rangle_2}$, respectively, where ${\left|\Theta\right\rangle_1=\lbrace\left|a_1\right\rangle,\left|a_2\right\rangle\rbrace}$ and ${\left|\Theta\right\rangle_2=\lbrace\left|a_1\right\rangle,\left|a_2\right\rangle,\left|a_3\right\rangle\rbrace}$:
\begin{align}
	&qm_1: qm_1(\left|a_1\right\rangle)=\sqrt{\psi_1},qm_1(\left|a_2\right\rangle)=\sqrt{(1-\psi_1)}e^{\rm{i}\theta_{12}}\notag\\
	&qm_2: qm_2(\left|a_1\right\rangle)=\sqrt{\psi_2},qm_2(\left|a_2\right\rangle)=\sqrt{(1-\psi_2)}\notag
\end{align}

In Example 3, the variations in $\psi_1$, $\psi_2$ and $\theta_{12}$ directly influence the changes in $qm_1$ and $qm_2$. Specially, $\psi_1$ is constrained within the interval $[0,1]$, while $\psi_2$ takes on predetermined values of 0.3, 0.5, 0.7 and 1. Both $qm_1$ and $qm_2$ exhibit the same subsets and support values on $2^{\left|\Theta\right\rangle_1}$ and $2^{\left|\Theta\right\rangle_2}$. Consequently, as we manipulate $\psi_1$, $\psi_2$ and $\theta_{12}$, the corresponding QCI on $2^{\left|\Theta\right\rangle_1}$ and $2^{\left|\Theta\right\rangle_2}$ are shown in Fig. \ref{Example3}.

Analysis of Fig. \ref{Example3}(a), (b), (c) and (d), QCI reveals that the variation from $2^{\left|\Theta\right\rangle_1}$ to $2^{\left|\Theta\right\rangle_2}$ has no impact on the QCI. Even as $\psi_1$, $\psi_2$ and $\theta_{12}$ change, the value of QCI remains constant in both $2^{\left|\Theta\right\rangle_1}$ and $2^{\left|\Theta\right\rangle_2}$. Moreover, Fig. \ref{Example3}(e), (f), (g) and (h) demonstrate that when we change the input $QCI(qm_1,qm_2)$ to $QCI(qm_2,qm_1)$, the results remain identical. This observation confirms the symmetry of the QCI.\\

\vspace{0.2em}\noindent\textbf{Example 4} Assuming that there are two QMFs, $qm_1$ and $qm_2$, on $2^{\left|\Theta\right\rangle}$:
\begin{align}
	qm_1: &qm_1(\left|a_1\right\rangle)=\frac{0.1+y\rm{i}}{\sqrt{0.82+2y^2}}, qm_1(\vartheta_\omega)=\frac{0.9+y\rm{i}}{\sqrt{0.82+2y^2}}\notag\\
	qm_2: &qm_2(\left|a_1\right\rangle)=\frac{0.9+y\rm{i}}{\sqrt{0.82+2y^2}}, qm_2(\vartheta_\omega)=\frac{0.1+y\rm{i}}{\sqrt{0.82+2y^2}}\notag
\end{align}

\begin{figure}[!h]
	\includegraphics[width=3in]{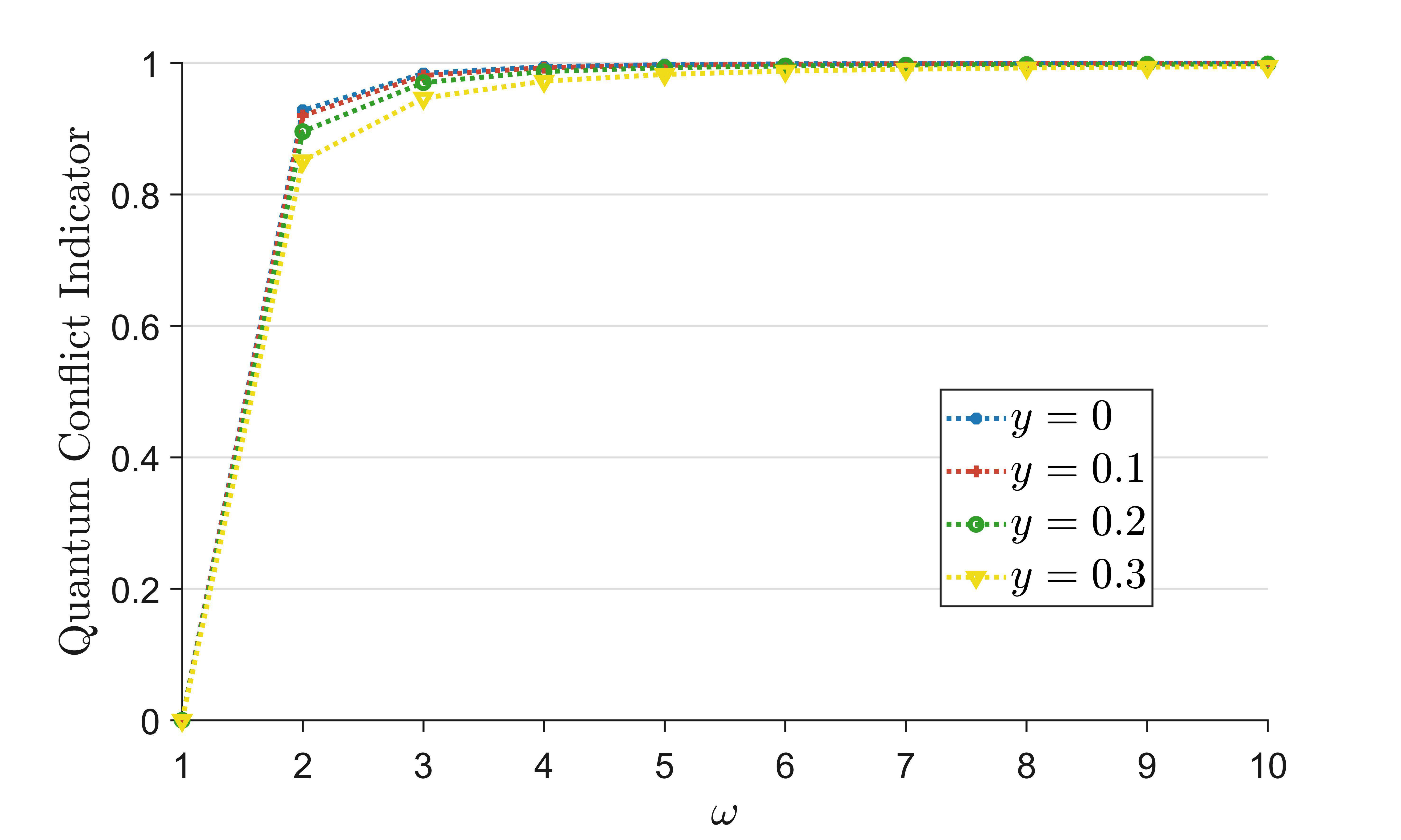}%
	\caption{The Quantum Conflict Indicator in Example 4.}
	\label{Example 5}
\end{figure}

Example 4 is obtained from \cite{pan2022distance}. Note that the angular phase is constrained within the interval $[0,\frac{\pi}{2}]$, and both the real and imaginary parts are positive. In this example, $qm_1$ and $qm_2$ vary based on the value of $y$ and the subset of $\vartheta_\omega$. Here $y$ takes on the predetermined values of  $0$, $0.1$, $0.2$ and $0.3$, while $\vartheta_\omega$ is set as shown in Table \ref{T1}. This example provides a comprehensive analysis of the changing behavior of the QCI with respect to variations in $\vartheta_\omega$ and the impact of the imaginary component on the QCI. The QCI between $qm_1$ and $qm_2$ is shown in Fig. \ref{Example 5}.

Fig. \ref{Example 5} shows that regardless of the changes in $y$, the QCI increases as the size of the subset $\vartheta_\omega$ grows. This result is both intuitive and reasonable. The increase in the number of irrelevant elements between $\left|a_1\right\rangle$ and $\vartheta_\omega$, i.e., the value of ${|\vartheta_\omega\cup \left|a_1\right\rangle|-|\vartheta_\omega\cap\left|a_1\right\rangle|}$, leads to a greater discrepancy between $qm_1$ and $qm_2$, thereby intensifying the level of conflict between them. Furthermore, comparing the value of QCI at different $y$ under the same $\omega$, we encounter four distinct cases:\\
1) when ${y=0}$, we have ${qm_1(\left|a_1\right\rangle)=qm_2(\vartheta_\omega)=\frac{1}{\sqrt{82}}}$ and ${qm_2(\left|a_1\right\rangle)=qm_1(\vartheta_\omega)=\frac{9}{\sqrt{82}}}$;\\
2) when ${y=0.1}$, we have $qm_1(\left|a_1\right\rangle)=qm_2(\vartheta_\omega)=\frac{1}{2\sqrt{21}}+\frac{1}{2\sqrt{21}}\rm{i}$ and $qm_2(\left|a_1\right\rangle)=qm_1(\vartheta_\omega)=\frac{9}{2\sqrt{21}}+\frac{1}{2\sqrt{21}}\rm{i}$;\\
3) when ${y=0.2}$, we have $qm_1(\left|a_1\right\rangle)=qm_2(\vartheta_\omega)=\frac{1}{3\sqrt{10}}+\frac{2}{3\sqrt{10}}\rm{i}$ and $qm_2(\left|a_1\right\rangle)=qm_1(\vartheta_\omega)=\frac{3}{10}+\frac{2}{3\sqrt{10}}\rm{i}$;\\
4) when ${y=0.3}$, we have $qm_1(\left|a_1\right\rangle)=qm_2(\vartheta_\omega)=\frac{1}{10}+\frac{3}{10}\rm{i}$ and $qm_2(\left|a_1\right\rangle)=qm_1(\vartheta_\omega)=\frac{9}{10}+\frac{3}{10}\rm{i}$;

It is evident that the value of QCI decreases as the imaginary part increases, indicating the significant impact of the imaginary component on the QCI. However, this trend diminishes with the increase on $\omega$. This phenomenon can be attributed to the fact that the QCI is more responsive to changes in the number of irrelevant elements between $\left|a_1\right\rangle$ and $\vartheta_\omega$.
\begin{table}[!h]
	\centering
	\caption{The variation of $\vartheta_\omega$ in Example 4\label{T1}}
	\renewcommand{\arraystretch}{1.05}
	\setlength{\tabcolsep}{8pt}
	\begin{tabular}{ll}
		\hline
		$\omega$ & $\vartheta_\omega$ \\
		\hline
		1 & $\left|a_1\right\rangle$\\
		2 & $\left|a_1\right\rangle\cup\left|a_2\right\rangle$\\
		3 & $\left|a_1\right\rangle\cup\left|a_2\right\rangle\cup\left|a_3\right\rangle$\\
		4 & $\left|a_1\right\rangle\cup\left|a_2\right\rangle\cup\left|a_3\right\rangle\cup\left|a_4\right\rangle$\\
		5 & $\left|a_1\right\rangle\cup\left|a_2\right\rangle\cup\left|a_3\right\rangle\cup\left|a_4\right\rangle\cup\left|a_5\right\rangle$\\
		6 & $\left|a_1\right\rangle\cup...\cup\left|a_6\right\rangle$\\
		7 & $\left|a_1\right\rangle\cup...\cup\left|a_6\right\rangle\cup\left|a_7\right\rangle$\\
		8 & $\left|a_1\right\rangle\cup...\cup\left|a_6\right\rangle\cup\left|a_7\right\rangle\cup\left|a_8\right\rangle$\\
		9 & $\left|a_1\right\rangle\cup...\cup\left|a_6\right\rangle\cup\left|a_7\right\rangle\cup\left|a_8\right\rangle\cup\left|a_9\right\rangle$\\
		10 & $\left|a_1\right\rangle\cup...\cup\left|a_6\right\rangle\cup\left|a_7\right\rangle\cup\left|a_8\right\rangle\cup\left|a_9\right\rangle\cup\left|a_{10}\right\rangle$\\
		\hline
	\end{tabular}
\end{table}

\section{QCI-based conflict fusion method}
In this section, the QCI is employed to the multi-source information fusion method introduced by Deng in \cite{pan2022distance}. The objective is to mitigate the influence of conflicting evidence on the decision-making process. Through experimental analysis, the results reveal that QCI exhibits certain advantages in this domain.
\subsection{Method describing}
For a set $qm$ with $n$ QMFs in ${2^{\left|\Theta\right\rangle}=\left\lbrace\emptyset,A_1,... ,A_{2^n-1}\right\rbrace}$, the specific steps of the QCI-based conflict fusion method are as follows:

\textbf{step 1:} For each ${qm_k, k=1,...,n}$, the QCI between it and every ${qm_i, i=1,...,n}$ are calculated by Eq (\ref{QCI}) to generate the QCI matrix $\mathcal{Q}$ as follow:
\begin{eqnarray}
	\begin{aligned}
		\mathcal{Q}=\left[\begin{matrix}QCI(qm_1,qm_1)&\cdots&QCI(qm_1,qm_n)\\\vdots&\ddots&\vdots\\\ QCI(qm_n,qm_1)&\cdots&QCI(qm_n,qm_n)\\\end{matrix}\right].
	\end{aligned}
\end{eqnarray}

\textbf{step 2:} The support degree for each QMF $qm_k, k=1,...,n$ in $qm$ is calculated as follow:
\begin{eqnarray}
	S_{qm_k}=\sum_{i=1,i\neq k}^{n}(1-QCI(qm_i,qm_k)).\label{step2}
\end{eqnarray}

\textbf{step 3:} The discount coefficient for each QMF $qm_k, k=1,...,n$ in $qm$ is calculated as follow:
\begin{eqnarray}
	D_{qm_k}=\sqrt{\frac{S_{qm_k}}{\sum_{n}^{i=1}S_{qm_i}}}.\label{step3}
\end{eqnarray}

\textbf{step 4:} The discount QMF for each QMF $qm_k, k=1,...,n$ in $qm$ is generated as follow:
	\begin{equation}
		DMF_k=\left\{
		\begin{aligned}
			&D_{qm_k}\times qm_k(A_j),A_j\in2^{\left|\Theta\right\rangle},A_j\neq\left|\Theta\right\rangle\\
			&\sum_{j=1}^{m}(1-D_{qm_k})\times qm_k(A_j)+qm_k(\left|\Theta\right\rangle),else
		\end{aligned}
		\right.\\ \label{step4}
	\end{equation}

\textbf{step 5:} After normalization of all discount QMFs, the discount QMFs are merged by Eq (\ref{DRC-QM}) as follows:
\begin{eqnarray}
	qm_{final}=((DMF_1\oplus DMF_2)\oplus...)\oplus DMF_n.\label{step5}
\end{eqnarray}

In order to better explain this method, its pseudo-code is shown in Algorithm \ref{alg1}.
\begin{algorithm}[!t]
	\small
	\caption{\begin{small}QCI-based conflict fusion method.\end{small}}
	\textbf{Input:} $QMFs$: a $n\times m$ matrix, where each of the $n$ rows corresponds to a QMF defined on the same QFoD. \\
	\textbf{Output:} $QMF_{final}$: the result of these $n$ QMFs after combining by DRC-QM.\\
		\begin{algorithmic}[1]
		\STATE\textit{for} i in [1,n]:
		\STATE\quad\textit{for} j in [1,n]:
		\STATE\quad\quad Calculation of the $QCI(qm_i,qm_j)$ by Eq (\ref{QCI});
		\STATE\quad\textit{end for}
		\STATE\textit{end for}
		\STATE\textit{for} i in [1,n]:
		\STATE\quad Calculation of the support degree by Eq (\ref{step2});
		\STATE\textit{end for}
		\STATE\textit{for} i in [1,n]:
		\STATE\quad Calculation of the discount coefficient by Eq (\ref{step3});
		\STATE\textit{end for}
		\STATE\textit{for} i in [1,n]:
		\STATE\quad Calculation of the discount QMF by Eq (\ref{step4});
		\STATE\textit{end for}
		\STATE Combination of the discount QMFs by Eq (\ref{step5});
	\end{algorithmic}
	\label{alg1}
\end{algorithm}

\subsection{Datasets and Benchmarks}
In this subsection, the QCI-based conflict fusion method is applied to various commonly utilized datasets from the UC Irvine Machine Learning Repository (UCI)\footnote{http://archive.ics.uci.edu/ml/index.php}, as outlined in Table \ref{UCI}. Subsequently, the superiority of the QCI is evaluated by comparing it against several fusion methods, namely DRC-QM \cite{pan2020a}, DRC \cite{shafer1992dempster}, the method proposed by Deng \cite{pan2022distance}, the method proposed by Murphy \cite{murphy2000combining}, the improved Proportional Conflict Redistribution Rule no.5 (PCR5+) \cite{dezert2021improvement} and the improved Proportional
Conflict Redistribution Rule no.6 (PCR6+) \cite{dezert2021improvement}.
 
\begin{table}[!h]
	\caption{Used datasets from UCI database\label{UCI}}
	\centering
	\renewcommand{\arraystretch}{1.05}
	\setlength{\tabcolsep}{10pt}
	\begin{tabular}{lccc}
		\hline
		Dataset & Numbers & Classes & Variables\\
		\hline
		Iris & 150 & 3 & 4\\
		Banknote &1372 &2 &4\\
		Transfusion &748 &2 &4\\
		Seed & 210 & 3 & 7\\
		Wholesale &440 &3 &7\\
		Breast Cancer & 699 & 2 & 9\\
		\hline
	\end{tabular}
\end{table}
\subsection{Experiment Settings and Results}
In this section, we select the experimental procedure presented by Deng in \cite{pan2022distance} and the method described in \cite{pan2020a} is employed to generate QMFs according to the following procedure:
\begin{eqnarray}
	qm_k(F_j^k)=(\frac{1}{e^{|\sigma_j^{k'}-\sigma_j^k|}}e^{\rm{i}|\mu_j^{k'}-\mu_j^k|}),\label{gQMF}
\end{eqnarray}
where $\sigma_j^k$ is the standard deviation of the training set corresponding to the $F_j^k$, while $\sigma_j^{k'}$ refers to the standard deviation of the combination of the training set and test set corresponding to the $F_j^k$. Similarly, $\mu_j^k$ is the mean of the training set corresponding to the $F_j^k$, while $\mu_j^{k'}$ refers to the mean of the combination of the training set and test set corresponding to the $F_j^k$.

\begin{figure*}[tp]
	\centering
	\subfloat[Iris]{\includegraphics[width=2.2in]{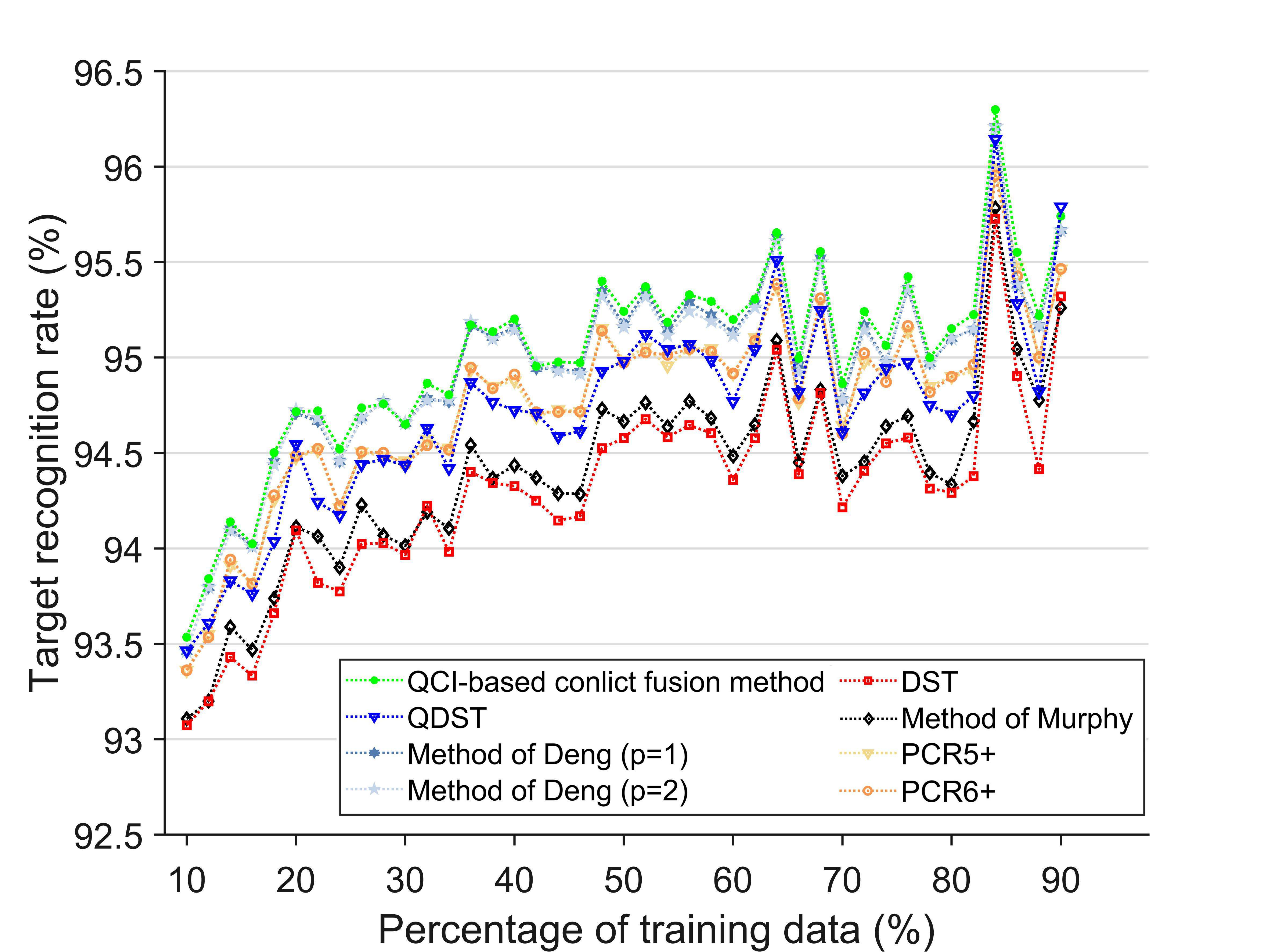}%
		\label{UCIresult1}}
	\hfil
	\subfloat[Banknote]{\includegraphics[width=2.2in]{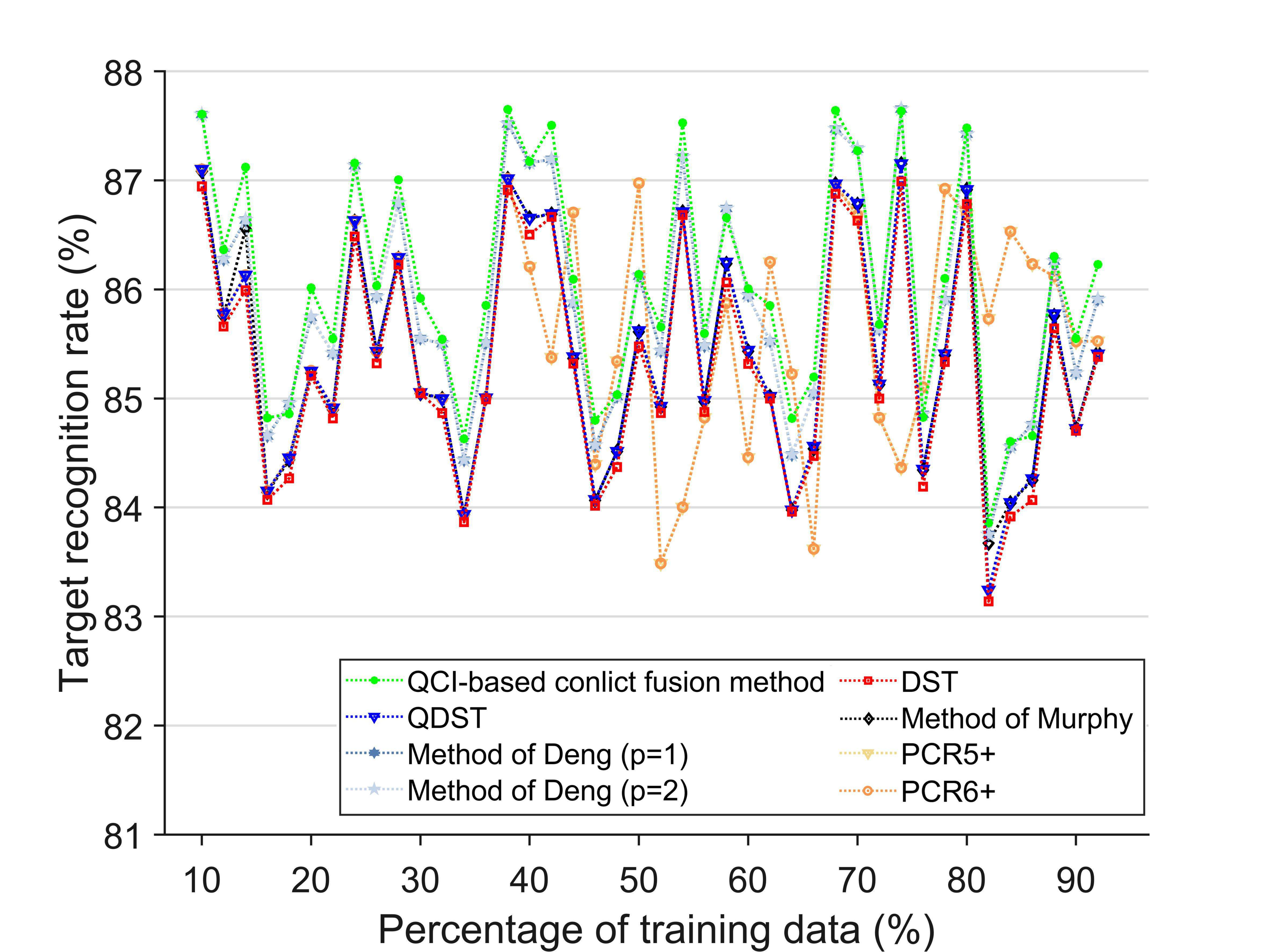}%
		\label{UCIresult2}}
	\hfil
	\subfloat[Transfusion]{\includegraphics[width=2.2in]{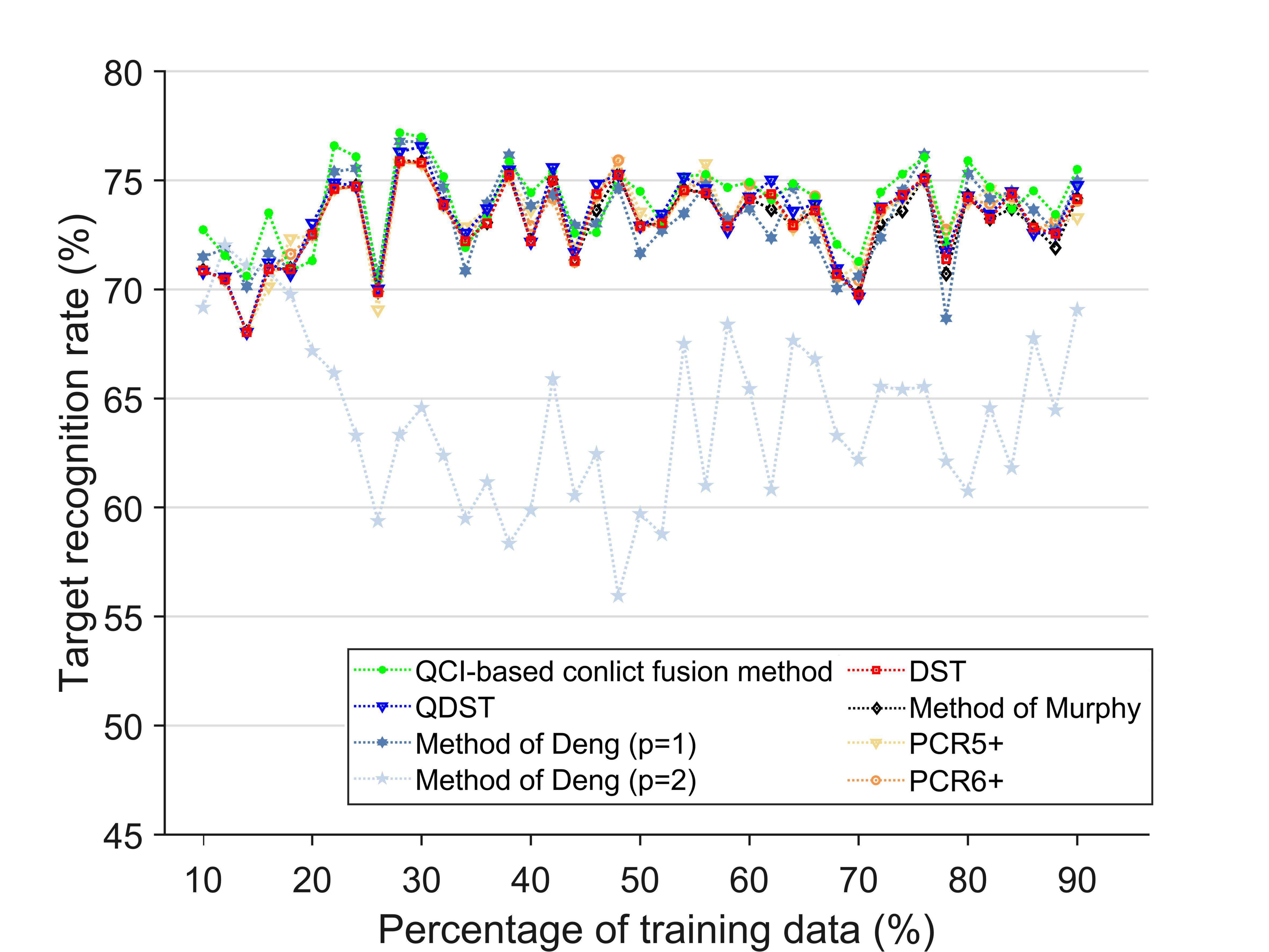}%
		\label{UCIresult3}}
	\hfil
	\subfloat[Seed]{\includegraphics[width=2.2in]{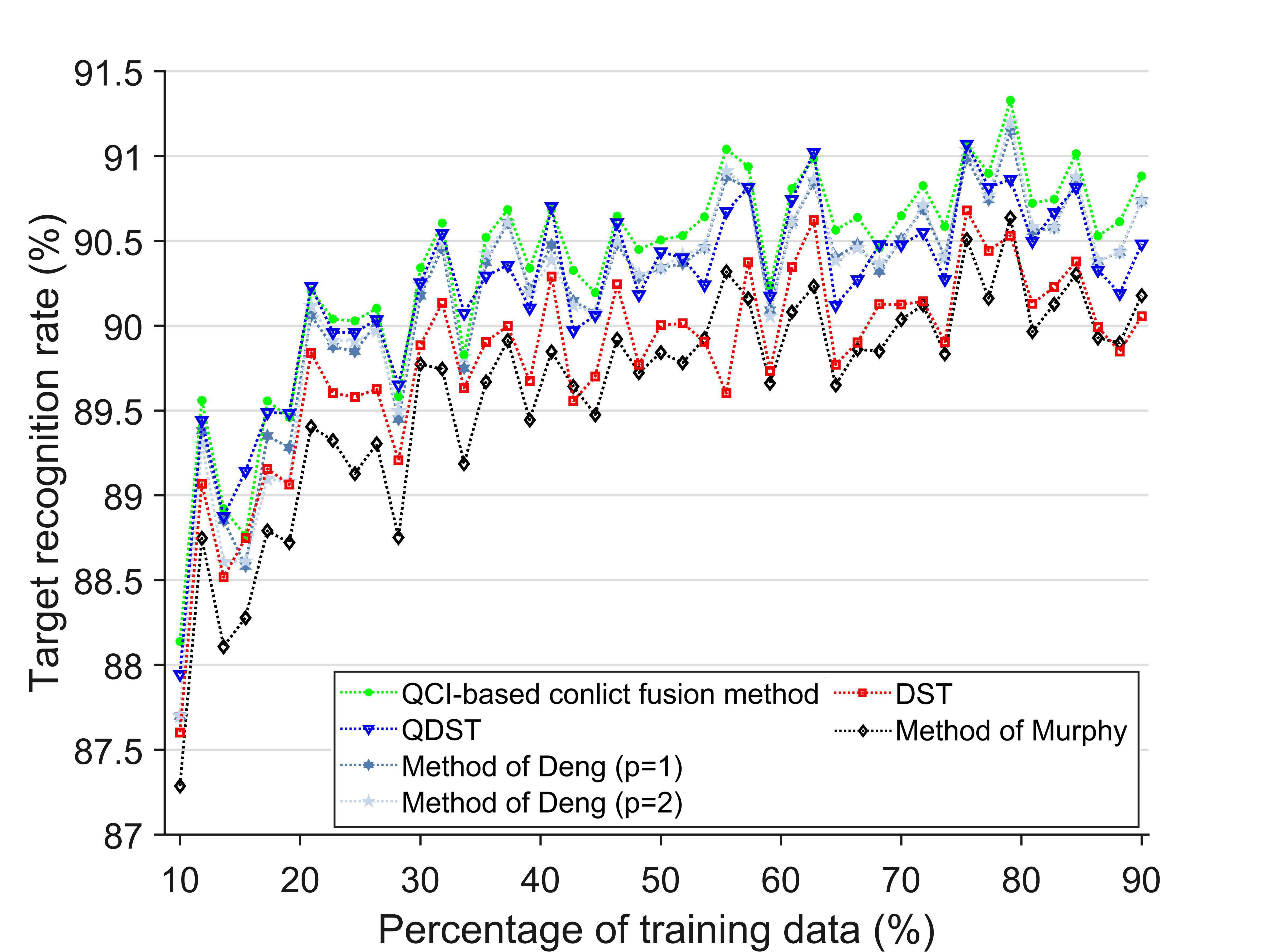}%
		\label{UCIresult4}}
	\hfil
	\subfloat[Wholesale]{\includegraphics[width=2.2in]{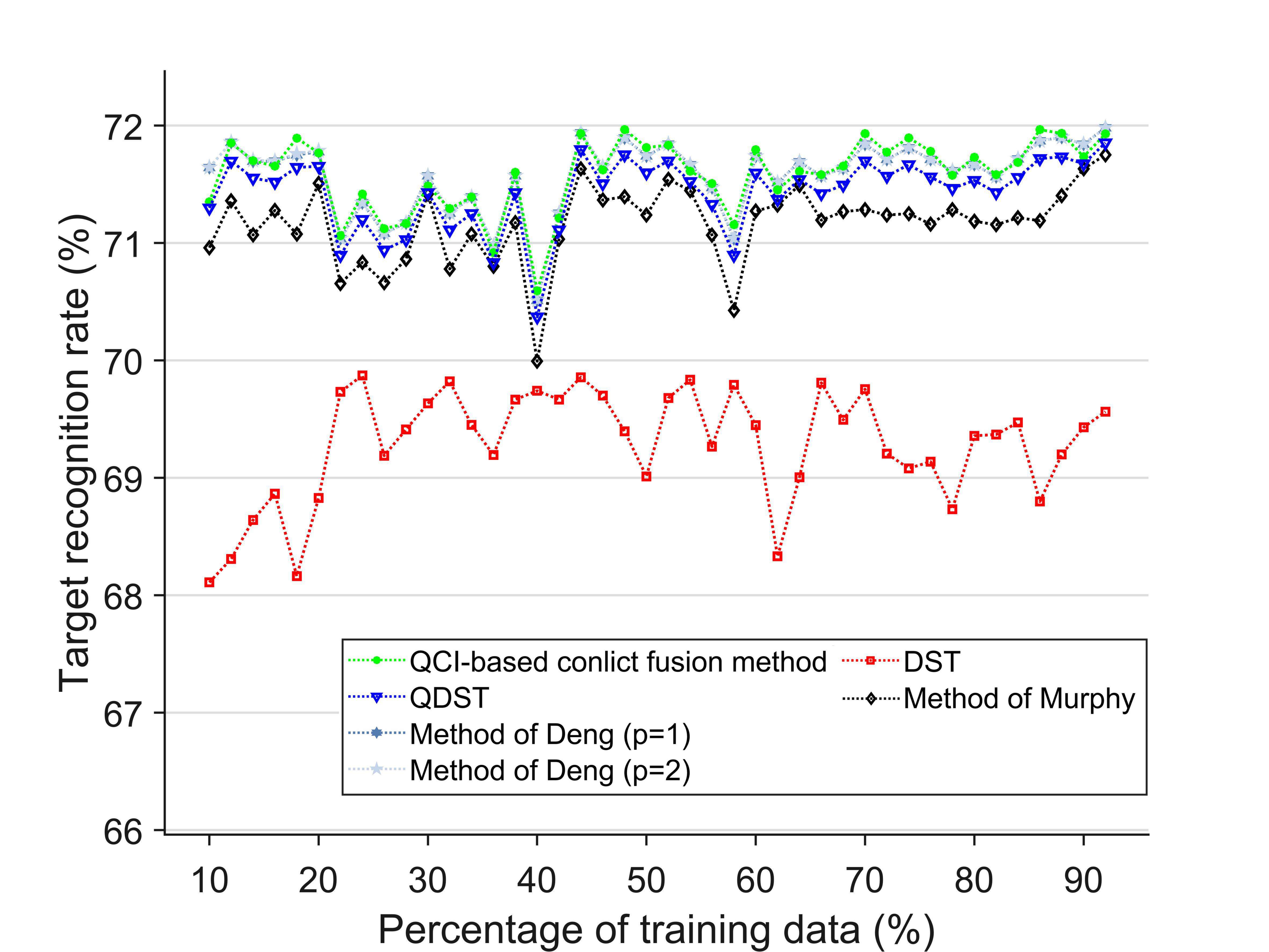}%
		\label{UCIresult6}}
	\hfil
	\subfloat[Breast Cancer]{\includegraphics[width=2.2in]{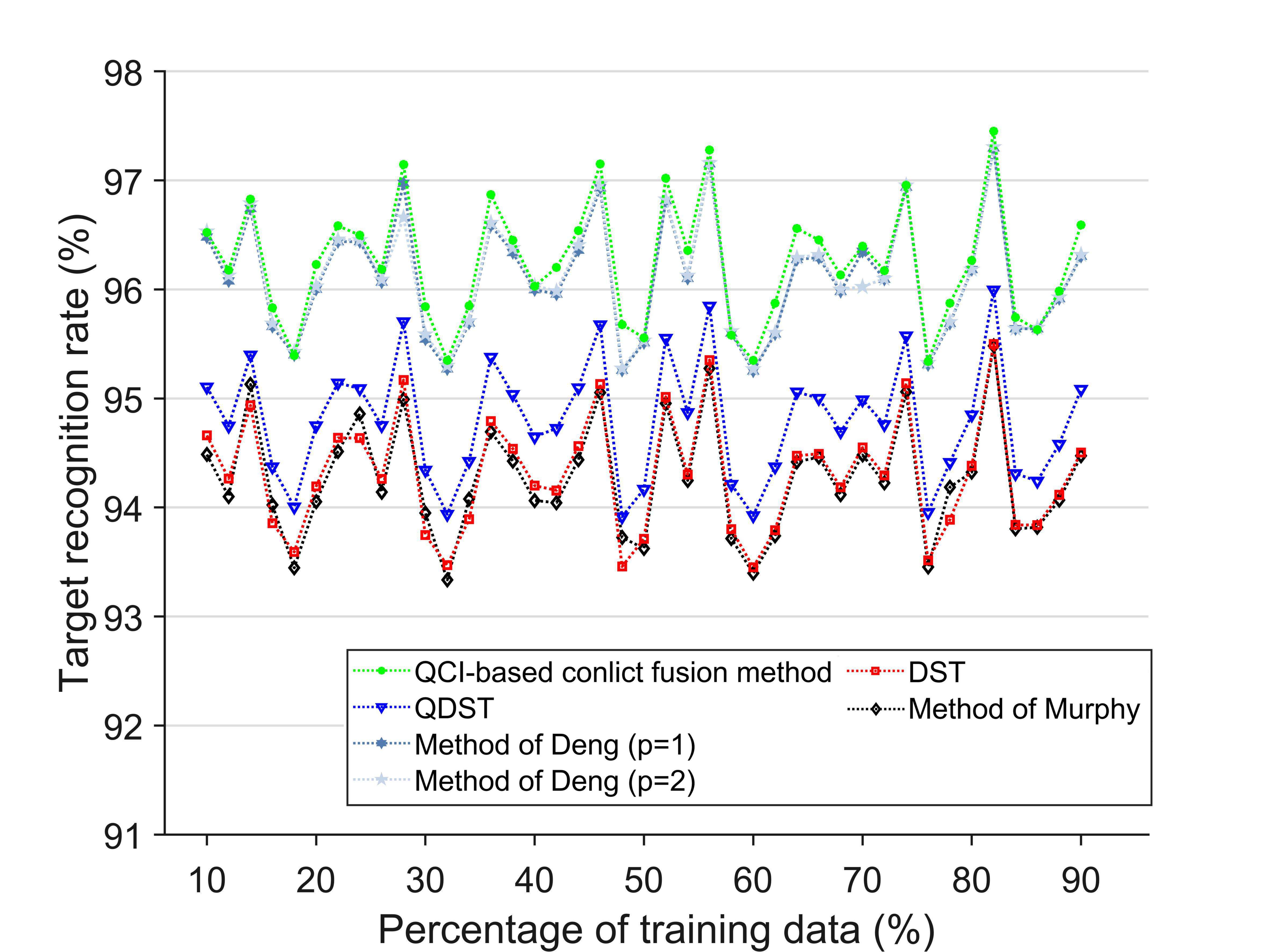}%
		\label{UCIresult5}}
	\hfil
	\caption{The accuracy of target recognition changes with the increase of the number of training set.}
	\label{UCIresult}
\end{figure*}

We perform one hundred Monte Carlo validation experiments for each scenario and used Pignistic Probability Transform \cite{pan2022distance} to obtain the final decision probabilities. The experimental results are presented in Fig. \ref{UCIresult}, which provide a assessment of the performance of the proposed fusion method. Please note that we exclusively employed the PCR5+ and PCR6+ techniques on Iris dataset, Banknote dataset and Transfusion dataset. Although these two methods demonstrated satisfactory performance across the aforementioned datasets, their algorithmic intricacy is significantly high. Consequently, when confronted with a substantial number of elements within the frame of discernment or an extensive body of evidence involved in the fusion process, such as Seed dataset, Wholesale dataset and Breast Cancer dataset, the associated time overhead becomes prohibitively burdensome, rendering result generation nearly unattainable. For detail, suppose there are $m$ mass functions defined on the power set of $\Theta$, which has $n$ elements. Subsequently, the algorithmic complexity of the PCR5+ method is $\rm{O(m^3\times n^m)}$, while the PCR6+ algorithmic complexity is $\rm{O(m^4\times n^m)}$. In comparison, the algorithmic complexities of the DST and QDST are $\rm{O(n^m)}$, whereas the complexity of our proposed QCI-based conflict fusion method corresponds to $max\lbrace\rm{O(n^m)},\rm{O(n^n)}\rbrace$.

The Fig. \ref{UCIresult}(a) and Fig. \ref{UCIresult}(d) presents the accuracy of unrecognized target recognition in the Iris dataset and Seed dataset, respectively. These figures reveal that the accuracy of all methods increases as the number of training sets grows. In addition, our fusion method largely outperforms other fusion methods. Meanwhile, when compared to the method of Deng, which is also a conflict detection method in QDST, our fusion method achieves higher accuracy, thus indicating its ability to better capture conflicts among QMFs. 

The Fig. \ref{UCIresult}(b), Fig. \ref{UCIresult}(c), Fig. \ref{UCIresult}(e) and Fig. \ref{UCIresult}(f) present the accuracy of unrecognized target recognition in the Banknote dataset, Transfusion dataset, Wholesale dataset and Breast Cancer dataset correspondingly. As observed from the figures, the accuracy of these methods fluctuates with the increase of the training set. However, our fusion method outperforms other methods on both datasets in most cases. Furthermore, in the case of the Wholesale dataset, characterized by notable data variance and volatility, the QMF-based fusion method surpasses conventional methodologies. This superiority stems from its consideration of these factors as angular phases during the quality function generation process. Additionally, in the context of the Transfusion dataset, the performance of the method of Deng at $p=2$ is bad, while the method of Deng at $p=1$ yields favorable outcomes. This observation aligns with the findings presented in \cite{pan2022distance}, emphasizing the significant impact of the choice of parameter $p$ on the effectiveness of the method of Deng. Nonetheless, our fusion method outperforms the method of Deng across the majority of data points. The performances further showcases the generalisability and superiority of our fusion method. To further explore the potential of QCI-based conflict fusion method, we extend its application to the intricate task of Out-of-Distribution (OOD) detection.

\section{Application of QCI in Out-of-Distribution Detection}
In this section, QCI-based conflict fusion method is referenced with OOD detection and put forth the Class Description Domain Space (C-DDS) and its optimized version, C-DDS+ to effectively capitalize on the capabilities of QCI in intricate scenarios.
\subsection{Base model: C-DDS}
\begin{figure*}[!h]
	\centering
	\includegraphics[width=6.5in]{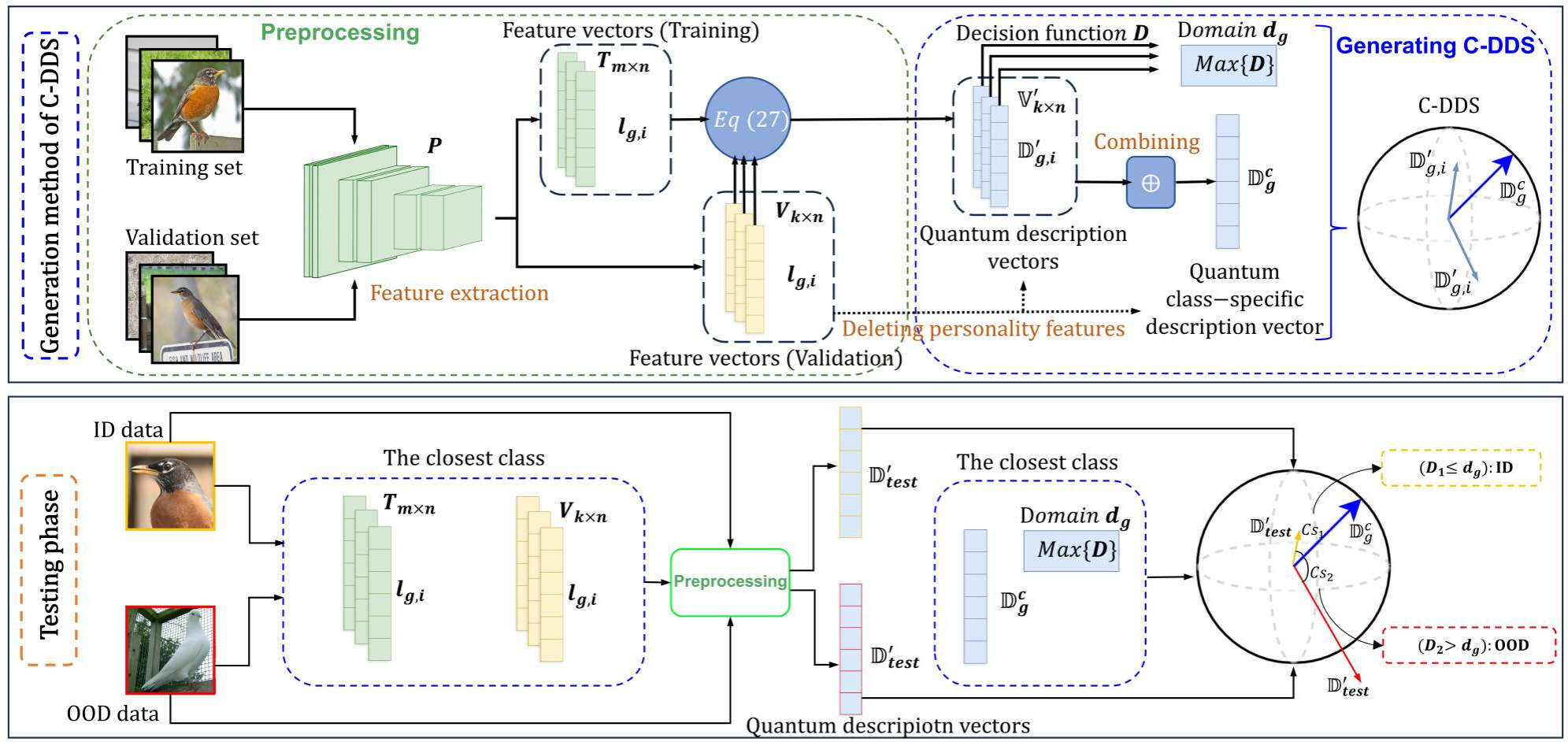}%
	\caption{The specific processes for C-DDS.}
	\label{C-DDS}
\end{figure*}

An expressive quantum description vector is generated for each In-Distribution (ID) data belonging to each ID class. However, the inherent diversity within the training dataset introduces the possibility of high conflict between certain ID data of the same ID class, leading to significant disparities in their respective quantum description vectors. To address this issue, we propose the C-DDS by utilizing the QCI-based fusion method. The step-by-step procedure is depicted in Fig. \ref{C-DDS}, and to enhance clarity, a pseudo-code representation is provided in Algorithm \ref{C-DDS_A}.

\begin{algorithm}[!t]
	\small
	\caption{\begin{small}Generating C-DDS and Testing phase.\end{small}}
	\textbf{Input:} $P$: a pretrained network, $T_{m\times n}$: the feature matrix of training data, $V_{k\times n}$: the feature matrix of validation data, $l_{test}$: the feature vector of test instance, $f$: the number of ID classes\\
	\textbf{Output:} $Score_c$: the score of the test instance\\
	\begin{algorithmic}[1]
		\STATE \textbf{function} \textsc{Preprocessing}
		\STATE \textit{for} i in [1,n]:
		\STATE \quad Calculating the quantum description vector $\mathbb{D}_{g,i}$ by Eq (\ref{gQMF});
		\STATE \textit{end for}
		\STATE Establishment of the quantum description matrix $\mathbb{V}_{k\times n}=[\mathbb{D}_{g,1},...,\mathbb{D}_{g,k}]^{\top}$;
		\STATE Calculation of the variance for each column of $V_{k\times n}$;
		\STATE Establishment of the individualistic features for $V_{k\times n}$;
		\STATE Establishment of the processed quantum description matrix $\mathbb{V}_{k\times n}'=[\mathbb{D}_{g,1}',...,\mathbb{D}_{g,k}']^{\top}$;
		\STATE \textbf{end function}
		\vspace{1em}
		\STATE \textbf{function} \textsc{Generating C-DDS}
		\STATE Generation of the quantum class-specific description vector $\mathbb{D}^c_g$ by Algorithm \ref{alg1};
		\STATE \textit{for} i in [1,k]:
		\STATE \quad Calculating the decision function $\mathit{D}(\mathbb{D}_{g,i}',\mathbb{D}^c_g)$ by Eq (\ref{cos});
		\STATE \textit{end for}
		\STATE Calculation of the domain $d_g$ by Eq (\ref{domain});
		\STATE \textbf{end function}
		\vspace{1em}
		\STATE \textbf{function} \textsc{Testing phase}
		\STATE Establishment of the predicted class $g$ by highest softmax probability of $l_{test}$;
		\STATE Importation of the $T_{m\times n}$, $V_{k\times n}$, $\mathbb{D}^c_g$ and $d_g$;
		\STATE Generation of the processed quantum description vector $\mathbb{D}_{test}'$ by \textsc{Preprocessing};
		\STATE Calculation of the decision function $\mathit{D}(\mathbb{D}_{test}',\mathbb{D}^c_g)$ by Eq (\ref{cos});
		\STATE Calculation of the $Score_c$ by Eq (\ref{scorec})
		\STATE \textbf{end function}
		\vspace{1em}
		\STATE \textit{for} g in [1,f]:
		\STATE\quad Generation of C-DDS by \textsc{Generating C-DDS};
		\STATE \textit{end for}
		\STATE Test of the test instance by \textsc{Testing phase};
	\end{algorithmic}
	\label{C-DDS_A}
\end{algorithm}

\textbf{Preprocessing}: For a pretrained network $P$, the feature vector $l\in\mathbb{R}^n$ of each ID data can be easily obtained through the feature extraction operation performed by $P$. Assuming that each $l$ consists of $n$ feature elements, the matrix comprising the training data of the same class $g$ as $T_{m\times n}=[l_{g,1},...,l_{g,m}]^{\top}\in\mathbb{R}^{m\times n}$ and the matrix comprising the validation data of the same class as $V_{k\times n}=[l_{g,1},...,l_{g,k}]^{\top}\in\mathbb{R}^{k\times n}$. In this context, $T_{m\times n}$ and $V_{k\times n}$ is utilized to generate a quantum description vector for each ID class employing the Eq (\ref{gQMF}). The matrix consisting of all the generated quantum description vectors is denoted as $\mathbb{V}_{k\times n}=[\mathbb{D}_{g,1},...,\mathbb{D}_{g,k}]^{\top}\in\mathbb{C}^{k\times n}$. Subsequently, the variance is calculated for each column of the matrix $V_{k\times n}$, which reflects the fluctuation of the corresponding feature. Since $l_{g,1},...,l_{g,k}$ originate from the same ID class, they should exhibit a high level of commonality, i.e., common features. Features with significant fluctuations, on the other hand, should be interpreted as individualistic features specific to a particular data, which could introduce biases in decision making and should thus be disregarded. The processed $\mathbb{V}_{k\times n}$ is denoted as $\mathbb{V}_{k\times n}'=[\mathbb{D}_{g,1}',...,\mathbb{D}_{g,k}']^{\top}\in\mathbb{C}^{k\times n}$. 

\textbf{Generating C-DDS}: Given the presence of $f$ classes of ID data, we proceed to fuse the processed quantum description vectors $\mathbb{D}_{g,i}'\in\mathbb{C}^n,g=1,...,f$ within each class by employing the fusion method described in Chapter 4. Here, each $\mathbb{D}_{g,i}'$ is treated as a individual QMF. This fusion process ultimately yields the generation of quantum class-specific description vectors $\mathbb{D}^c_g$ for all ID classes. Subsequently, it becomes necessary to determine the corresponding domain for each ID class. To begin with, within a single class $g$, the decision function between each description vector and the quantum class-specific description vector is computed as follows:
\begin{eqnarray}
	\mathit{D}(\mathbb{D}_{g,i}',\mathbb{D}^c_g) = \frac{[\ \ \overline{\mathbb{D}_{g,i}'}(\mathbb{D}^c_g)^{\top}\ ]^2}{\overline{\mathbb{D}_{g,i}'}(\mathbb{D}_{g,i}')^{\top}\times\overline{\mathbb{D}^c_g}(\mathbb{D}^c_g)^{\top}},\label{cos}
\end{eqnarray}
where $\overline{\mathbb{D}_{g,i}'}$ and $\overline{\mathbb{D}^c_g}$ represent the conjugates of $\mathbb{D}_{g,i}'$ and $\mathbb{D}^c_g$, respectively. $(\mathbb{D}_{g,i}')^{\top}$ and $(\mathbb{D}^c_g)^{\top}$ indicate the transpositions of $\mathbb{D}_{g,i}'$ and $\mathbb{D}^c_g$, respectively. Meanwhile, $\mathbb{D}^c_g$ is the quantum class-specific description vector of ID class $g$ and the domain associated with each ID class is defined as follow:
\begin{eqnarray}
	\mathit{d_g}= max\lbrace\mathit{D}(\mathbb{D}_{g,1}',\mathbb{D}^c_g),...,\mathit{D}(\mathbb{D}_{g,k}',\mathbb{D}^c_g)\rbrace.\label{domain}
\end{eqnarray}

The set $\lbrace\mathit{D}(\mathbb{D}_{g,1}',\mathbb{D}^c_g),...,\mathit{D}(\mathbb{D}_{g,k}',\mathbb{D}^c_g)\rbrace$ consists of the plausible decision functions. By obtaining the domain and the corresponding quantum class-specific description vector for each class, it is possible to identify the C-DDS for each class.

\textbf{Testing phase}: To generate the processed quantum description vector $\mathbb{D}_{test}'\in\mathbb{C}^n$ for a test instance, a specific ID class is typically depended. However, this approach incurs a significant overhead at test time. To mitigate this, we adopt the strategy presented in the cited reference \cite{lee2018simple}, where the predicted class is selected by the highest softmax probability. This predicted class is then compared only to the C-DDS of that class, reducing testing overhead and increasing speed.

Let's assume that the predicted class for the test instance is denoted as the category $g$. Therefore, there are the feature matrix of its training data $T_{m\times n}$, the feature matrix of its validation data $V_{k\times n}$, its quantum class-specific description vector $\mathbb{D}^c_g$ and its domain $\mathit{d_g}$. Given $T_{m\times n}$, $V_{k\times n}$ and the feature vector of the test data $V_{test}$, the processed quantum description vector $\mathbb{D}_{test}'$ can be generated for this test instance. By calculating the decision function between $\mathbb{D}_{test}'$ and $\mathbb{D}^c_g$ using Eq (\ref{cos}), it is possible to discriminate between ID and OOD class as follows:
\begin{eqnarray} 
	\begin{cases}
		ID\ class & \text{if } D(\mathbb{D}_{test}'$, $\mathbb{D}^c_g)\leq\mathit{d_g} \\
		OOD\ class & \text{if } else
	\end{cases}.
\end{eqnarray}

To accommodate various scenario requirements, it is often desirable to assign a score to each test data. In the case of the C-DDS method, the score $Score_c$ for a test instance can be defined as follows:
\begin{eqnarray} 
	Score_c(\mathbb{D}_{test}',\mathbb{D}^c_g) = \frac{\mathbb{D}_{test}'}{\mathbb{D}^c_g}.\label{scorec}
\end{eqnarray}
Note that the larger the $Score_c$ of the test instance, the more likely it is to be classified as an OOD class.

\subsection{Improved model: C-DDS+}
In the context of OOD detection, the C-DDS method integrates the softmax probability and the feature information, but it does not take into account the presence of the logit of the pretrained network. To further optimize our OOD detection method and address the vast diversity of test instances, we propose leveraging the Proportional
Conflict Redistribution Rule no.5 (PCR5)\cite{dezert2021improvement} for combining information.

The Decoupling MaxLogit method (DML), introduced by Zhang and Xiang in \cite{zhang2023decoupling}, distinguishes itself from other OOD detection methods that solely rely on logit information, such as ODIN \cite{liang2018enhancing} and MaxLogit \cite{basart2022scaling}. It considers both the available information between the logit and the fully connected layer of the pretrained network, as well as the information embedded in the logit itself. 

In our enhanced model (C-DDS+), the C-DDS combines the DML and C-DDS using DRC-GM. By simultaneously considering the features, softmax probabilities, and logits, C-DDS+ can improve the sensitivity and accuracy of OOD detection.

\textbf{Definition of PCR5:} For two mass functions, $m_1$ and $m_2$, on $2^{\Theta}=\left\lbrace\emptyset,X_1,... ,X_{2^n-1}\right\rbrace$ the PCR5 is defined as follows:\par
\begin{align}
&m(X_k)=\sum_{X_i\cap X_j=X_k}m_1(X_i)\cdot m_2(X_j)+\notag\\
&\sum\limits_{X_i\cap X_j=\emptyset\ and\ i\neq j}\frac{m_1(X_i)^2m_2(X_j)}{m_1(X_i)+m_2(X_j)}+\frac{m_2(X_i)^2m_1(X_j)}{m_2(X_i)+m_1(X_j)},\label{PCR5}
\end{align}

\textbf{Definition of DML:} In DML, the score $Score_d$ is defined as follows:
\begin{eqnarray} 
	Score_d = \lambda max\lbrace Cs(L,w_1),...,Cs(L,w_f)\rbrace + ||L||,
\end{eqnarray}
where $w_i\in\mathbb{R}^f,i=1,...,n$ represents a row vector of weight matrix $W=[w_1,...,w_n]^\top\in\mathbb{R}^{n\times f}$, $\lambda$ represents a hyper-parameter based on Gaussian noise and $||L||$ denotes the number of paradigms of the logit $L$. Furthermore, the cosine similarity between $L$ and $w_i$, $Cs(L,w_i)$, is defined as:
\begin{eqnarray} 
	Cs(L,w_i)=\frac{L^\top w_i}{||L||\cdot||w_i||}.
\end{eqnarray}
Note that the larger the $Score_d$ of the test instance, the more likely it is to be classified as an OOD class.

\textbf{Testing phase:} For a test instance, we obtain $Score_c=s_1$ in C-DDS and $Score_d=s_2$ in DML. If both C-DDS and DML are deemed reputable sources, denote them as $m_1$ and $m_2$, respectively. Regard the weights associated with OOD data as the measure of uncertainty within the discriminant framework $2^{\Theta}$ for the respective credible source. Note that since it is possible for both $s_1$ and $s_2$ to have values greater than $1$, scaling is performed before fusing these scores.
\begin{eqnarray}
	\begin{cases}
		s_1' = \mathit{k}_1+s_1\\
		s_2' = \mathit{k}_2+s_2
	\end{cases},
\end{eqnarray}
where $\mathit{k}_1$ represents a hyper-parameter that determines the scaling of $s_1$ and $\mathit{k}_2$ denotes a hyper-parameter that determines the scaling of $s_2$. Therefore, we have $m_1(\Theta)=s_1'$ and $m_2(\Theta)=s_2'$. According to Eq (\ref{PCR5}), the support degree for test instance outside of the classes represented by $\Theta$ after combining $m_1(\Theta)$ and $m_2(\Theta)$ can be represented as:
\begin{eqnarray}
	m(\Theta)=s_1'\times s_2'.
\end{eqnarray}
Thus, the composite score obtained by considering the features, softmax probabilities, and logits is given by $m(\Theta)$.

\section{Experiments}
In this section, a comparative analysis of our algorithm against the state-of-the-art OOD detection methods is conducted. For evaluation purposes, the mini-ImageNet dataset \cite{vinyals2016matching}, a subset of Imagenet-1k \cite{deng2009imagenet}, is used as the In-Distribution (ID) data. The performance of the methods is evaluated using both CNN-based models and Transformer-based models. A detailed description of the experimental setup employed in this study is presented below.

\subsection{Experiment Setup}
\textbf{OOD Datasets:} To comprehensively benchmark the algorithms, five OOD datasets are utilized \cite{huang2021mos}: 1) iNaturalist is a fine-grained species classification dataset. 2) Texture contains natural texture images. 3) ImageNet-O is a dataset specifically designed to test the robustness of visual models to OOD samples. 4) Sun contains images of different scenes. 5) Places is a dataset designed according to the principles of human visual cognition. 

\textbf{Evaluation Metrics:}Two commonly used metrics are reported. AUC \cite{9847099} is a threshold-free metric that computes the area under the receiver operating characteristic curve. Higher value indicates better detection performance. FPR95 \cite{9847099} is short for FPR@TPR95, which is the false positive rate when the true positive rate is 95\%. The smaller FPR95 the better.

\textbf{Experiment Settings:} We conduct experiments on two CNN-based models, WRN \cite{Zagoruyko2016WRN} and VGG \cite{kim2016accurate}, and two Transformer-based models, ViT \cite{dosovitskiy2018VIT} and Swin \cite{liu2021swin}. The pre-trained VGG16, WideResNet50-2, ViT-B-16, and Swin-B models from Imagenet-1K are utilized, and the models are fine-tuned on the mini-ImageNet dataset with a learning rate of 0.001, using the weights of the first hundred layers. The pre-training weights of all models are obtained from the PyTorch Vision library\footnote{https://pytorch.org/vision/stable/models.html}. It should be noted that the scaling hyper-parameter $k_1$ is always $0$ and the scaling hyper-parameter $k_2$ is set as $VGG:k_2=-82$ , $WRN:k_2=-14$, $VIT:k_2=-91$, $Swin:k_2=-1.6$, respectively,

\textbf{Baseline Methods:} Our method is compared with eight baselines that do not require fine-tuning. They are MSP \cite{dan2018baseline}, Energy \cite{liu2020energy}, MaxLogit \cite{basart2022scaling}, ODIN \cite{liang2018enhancing}, KL Matching \cite{basart2022scaling}, Mahalanobis \cite{lee2018simple}, VIM \cite{wang2022vim} and DML \cite{zhang2023decoupling}. For DML, the hyper-parameter is set as $VGG:\lambda=7.6$ , $WRN:\lambda=0.4$, $VIT:\lambda=8.6$, $Swin:\lambda=0.8$, respectively, and for all other methods, the presets from \cite{kirchheim2022pytorch}, a python library integrating these methods, are used.

\subsection{Results of comparative experiments}
We present the outcomes of the VGG model in Table \ref{VGG} and the WRN model in Table \ref{WRN}. Additionally, the outcomes of the Swin model are displayed in Table \ref{Swin} and the ViT model in Table \ref{ViT}. The best AUC and FPR95 are highlighted in bold, while the second-best results are underlined.\\
\textbf{Application of our method on CNN-based models:} Table \ref{VGG} presents the results of C-DDS and C-DDS+ with the VGG model on the iNaturalist, Texture, ImageNet-O, Sun, and Places datasets, respectively. C-DDS achieves the highest AUC and the lowest FPR95 among all approaches after C-DDS+ on the iNaturalist dataset and ImageNet-O. On average, C-DDS achieves an AUC of 92.19\%, which is the highest AUC after C-DDS+, surpassing the current state-of-the-art ViM method and the DML methods by 1.21\% and 0.61\%, respectively. This indicates the effectiveness of our proposed C-DDS method, which utilizes the QCI-based conflict fusion method to combine common class information and exclude individual characteristics, resulting in a more comprehensive feature representation of classes. However, it is worth noting that the average FPR95 of C-DDS reaches 30.51\%, which is 0.89\% lower than that of the DML method. Notably, on the Places dataset, the FPR95 of C-DDS is 6.33\% lower than that of DML, suggesting misclassification of more ID classes as OOD classes. We speculate that this can be attributed to the relatively weaker performance of the VGG16 model itself on the ID dataset compared to other models, leading to misclassifications of ID images. On the other hand, the mean value of the domains in the generated class description space for VGG is significantly lower compared to other models, indicating less variability among the feature vectors generated by the VGG model, and potentially resulting in too many common features masking certain effective features. Furthermore, C-DDS+ achieves the highest AUC and the lowest FPR95 across all datasets, suggesting that the fusion of C-DDS and DML effectively extracts ID information for OOD detection from features, softmax probabilities, and logits. Nevertheless, the excessively high FPR95 of C-DDS on the Places dataset indicates some misjudgments in the fusion process, leading to a 1.69\% higher FPR95 than DML. This issue can be mitigated by enhancing model accuracy and improving the expressiveness of extracted features.

Table \ref{WRN} presents the outcomes of C-DDS and C-DDS+ approaches using baseline methods with the WRN model on the respective datasets. C-DDS achieves the best AUC and FPR95 on the iNaturalist, ImageNet-O, and Texture datasets following C-DDS+. On average, C-DDS achieves an AUC of 92.19\%, surpassing the latest ViM and DML methods by 0.83\% and 0.19\%, respectively, making it the highest AUC after C-DDS+. Moreover, the average FPR95 of C-DDS is 13\%, second only to the low FPR95 of C-DDS+, and it is 5.03\% and 3.97\% lower than the current latest ViM method and DML method, respectively. Specifically, in the case of WRN, C-DDS+ achieves the highest AUC and the lowest FPR95 across all datasets. Compared to the original C-DDS and the latest ViM and DML methods, C-DDS+ improves the average AUC by 0.77\%, 0.96\%, and 1.63\%, while reducing the average FPR95 by 1.93\%, 5.9\%, and 6.96\%, respectively. These enhancements demonstrate the effectiveness of our method. Furthermore, from Tables \ref{VGG} and \ref{WRN}, our method exhibits greater advantages in models with higher accuracy and greater feature variability.\\
\textbf{Application of our method on Transformer-based models:} Table \ref{ViT} showcases the performance of C-DDS and C-DDS+ under the ViT model, while Table \ref{Swin} presents the performance under the Swin model. We observe that C-DDS and C-DDS+ achieve the best and second-best average AUC and FPR95 under these two models. Notably, the performance of C-DDS and C-DDS+ under the Transformer-based models outperforms that under the CNN-based models. This can be attributed to the inclusion of the self-attention in Transformer-based models, which enables effective information exchange between different image locations through global attentional interactions, resulting in the extraction of richer and more comprehensive feature representations. These findings suggest that our method is highly responsive to the features extracted by the model, and the presence of expressive features can further amplify the advantages of our proposed approach.

\begin{table*}[htbp]
	\centering
	\renewcommand{\arraystretch}{1.15}
	\setlength{\tabcolsep}{4.5pt}
	\caption{OOD detection of our proposed methods and the baseline methods under the VGG16 model with $Acc@1 \cite{9847099}=89.76\%$}
	\begin{tabular}{clrrrrrrrrrr|rr}
		\toprule
		\multirow{2}[2]{*}{Model} & \multicolumn{1}{c}{\multirow{2}[2]{*}{Method}} & \multicolumn{2}{c}{iNaturalist} & \multicolumn{2}{c}{Texture} & \multicolumn{2}{c}{ImageNet-O} & \multicolumn{2}{c}{Sun} & \multicolumn{2}{c|}{Places} & \multicolumn{2}{c}{average} \\
		&       & \multicolumn{1}{c}{AUC↑} & \multicolumn{1}{c}{FPR95↓} & \multicolumn{1}{c}{AUC↑} & \multicolumn{1}{c}{FPR95↓} & \multicolumn{1}{c}{AUC↑} & \multicolumn{1}{c}{FPR95↓} & \multicolumn{1}{c}{AUC↑} & \multicolumn{1}{c}{FPR95↓} & \multicolumn{1}{c}{AUC↑} & \multicolumn{1}{c|}{FPR95↓} & \multicolumn{1}{c}{AUC↑} & \multicolumn{1}{c}{FPR95↓} \\
		\midrule
		\multirow{10}[4]{*}{VGG} & MSP   & 89.63\% & 40.68\% & 86.12\% & 51.47\% & 56.01\% & 84.01\% & 81.63\% & 59.17\% & 81.04\% & 58.13\% & 78.89\% & 58.69\% \\
		& Energy & 92.49\% & 28.11\% & 92.05\% & 34.73\% & 64.55\% & 75.66\% & 87.75\% & 44.99\% & 84.99\% & 51.64\% & 84.37\% & 47.03\% \\
		& Maxlogit & 92.7\%  & 27.72\% & 91.78\% & 34.91\% & 64.15\% & 75.69\% & 87.49\% & 45.08\% & 85\%  & 51.16\% & 84.22\% & 46.91\% \\
		& ODIN  & 89.13\% & 37.75\% & 88.52\% & 40.38\% & 82.52\% & 56.38\% & 85.63\% & 43.71\% & 84.99\% & 45.55\% & 86.16\% & 44.75\% \\
		& KLMatching & 92.93\% & 27.56\% & 90.93\% & 39.58\% & 84.37\% & 47.5\%  & 84.24\% & 45.05\% & 82.83\% & 54.96\% & 87.06\% & 42.93\% \\
		& Mahalanobis & 91.74\% & 29.63\% & 91.83\% & 32.43\% & 84.59\% & 51.11\% & 89.15\% & 35.89\% & 90.14\% & 35.88\% & 89.49\% & 36.99\% \\
		& ViM   & 92.28\% & 23.32\% & \underline{93.1\%}  & \underline{22.81\%} & 91.72\% & 33.83\% & 88.52\% & 38.15\% & 89.12\% & 41.43\% & 90.98\% & 31.91\% \\
		& DML   & 93.85\% & 20.72\% & 92.75\% & 22.23\% & 90.3\%  & 36.93\% & \underline{90.75}\% & \underline{35.3\%}  & \underline{90.25\%} & \textbf{32.93\%} & 91.58\% & \underline{29.62\%} \\
		\cmidrule{2-14}          & C-DDS & \underline{95.43\%} & \underline{18.63\%} & 92.9\%  & 23.27\% & \underline{92.38\%} & \underline{32.96\%} & 90.31\% & 38.43\% & 89.91\% & 39.26\% & \underline{92.19\%} & 30.51\% \\
		& C-DDS+ & \textbf{96.33\%} & \textbf{15.06\%} & \textbf{94.66\%} & \textbf{17.25\%} & \textbf{93.18\%} & \textbf{30.57\%} & \textbf{91.74\%} & \textbf{33.5\%} & \textbf{92.01\%} & \underline{34.62\%} & \textbf{93.58\%} & \textbf{26.2\%} \\
		\bottomrule
	\end{tabular}%
	\label{VGG}%
	\vspace{1.5em}
	\centering
	\renewcommand{\arraystretch}{1.15}
	\setlength{\tabcolsep}{4.5pt}
	\caption{OOD detection of our proposed methods and the baseline methods under the WideResNet50-2 model with $Acc@1 \cite{9847099}=95.84\%$}
	\begin{tabular}{clrrrrrrrrrr|rr}
		\toprule
		\multirow{2}[2]{*}{Model} & \multicolumn{1}{c}{\multirow{2}[2]{*}{Method}} & \multicolumn{2}{c}{iNaturalist} & \multicolumn{2}{c}{Texture} & \multicolumn{2}{c}{ImageNet-O} & \multicolumn{2}{c}{Sun} & \multicolumn{2}{c|}{Places} & \multicolumn{2}{c}{average} \\
		&       & \multicolumn{1}{c}{AUC↑} & \multicolumn{1}{c}{FPR95↓} & \multicolumn{1}{c}{AUC↑} & \multicolumn{1}{c}{FPR95↓} & \multicolumn{1}{c}{AUC↑} & \multicolumn{1}{c}{FPR95↓} & \multicolumn{1}{c}{AUC↑} & \multicolumn{1}{c}{FPR95↓} & \multicolumn{1}{c}{AUC↑} & \multicolumn{1}{c|}{FPR95↓} & \multicolumn{1}{c}{AUC↑} & \multicolumn{1}{c}{FPR95↓} \\
		\midrule
		\multirow{10}[4]{*}{WRN} & MSP   & 95.01\% & 24.92\% & 89.25\% & 48.7\%  & 63.83\% & 84.25\% & 89.62\% & 45.84\% & 88.99\% & 46.24\% & 85.34\% & 49.99\% \\
		& Energy & 95.3\%  & 18.91\% & 90.7\%  & 45.94\% & 71.01\% & 81.17\% & 91.04\% & 39.65\% & 89.71\% & 45.2\%  & 87.55\% & 46.17\% \\
		& Maxlogit & 95.95\% & 18.19\% & 91.11\% & 45.24\% & 74.85\% & 77.65\% & 91.46\% & 39\%    & 90.21\% & 44.83\% & 88.72\% & 44.98\% \\
		& ODIN  & 95.5\%  & 18.63\% & 91.99\% & 45.96\% & 74.02\% & 79.57\% & 90.8\%  & 39.08\% & 89.32\% & 46.11\% & 88.32\% & 45.87\% \\
		& KLMatching & 97.38\%  & 12.6\%  & 95\%    & 38\%    & 80.66\% & 72.8\%  & 92.19\% & 31.09\% & 91.6\%  & 38\%    & 91.37\% & 38.5\% \\
		& Mahalanobis & 96.79\% & 17.31\% & 97.02\% & 12.53\% & 93.91\% & 25.51\% & 92.9\%  & 30.56\% & 92.03\% & 34.04\% & 94.53\% & 23.99\% \\
		& ViM   & 96.94\% & 13.56\%  & \underline{98.39\%} & 7.26\%  & 95.61\% & 25.99\% & 94.59\% & 21.29\% & 94.13\% & 22.03\% & 95.93\% & 18.03\% \\
		& DML   & 98.08\% & 7.17\%  & 97.22\% & 12.68\% & 94.29\% & 34.66\% & \underline{96.88\%} & \underline{13.43\%} & \underline{96.51\%} & \underline{16.93\%} & 96.6\% & 16.97\% \\
		\cmidrule{2-14}          & C-DDS & \underline{98.49\%} & \underline{5.1\%}   & 98.29\% & \underline{5.8\%}   & \underline{95.88\%} & \underline{19.15\%} & 95.83\% & 16.41\% & 95.46\% & 18.53\% & \underline{96.79\%} & \underline{13\%} \\
		& C-DDS+ & \textbf{98.95\%} & \textbf{3.61\%} & \textbf{98.65\%} & \textbf{4.74\%} & \textbf{96.43\%} & \textbf{19.56\%} & \textbf{97.07\%} & \textbf{12.63\%} & \textbf{96.67\%} & \textbf{14.84\%} & \textbf{97.56\%} & \textbf{11.07\%} \\
		\bottomrule
	\end{tabular}%
	\label{WRN}%
	\vspace{1.5em}
	\centering
	\renewcommand{\arraystretch}{1.15}
	\setlength{\tabcolsep}{4.5pt}
	\caption{OOD detection of our proposed methods and the baseline methods under the ViT-B-16 model with $Acc@1 \cite{9847099}=93.25\%$}
	\begin{tabular}{clrrrrrrrrrr|rr}
		\toprule
		\multirow{2}[2]{*}{Model} & \multicolumn{1}{c}{\multirow{2}[2]{*}{Method}} & \multicolumn{2}{c}{iNaturalist} & \multicolumn{2}{c}{Texture} & \multicolumn{2}{c}{ImageNet-O} & \multicolumn{2}{c}{Sun} & \multicolumn{2}{c|}{Places} & \multicolumn{2}{c}{average} \\
		&       & \multicolumn{1}{c}{AUC↑} & \multicolumn{1}{c}{FPR95↓} & \multicolumn{1}{c}{AUC↑} & \multicolumn{1}{c}{FPR95↓} & \multicolumn{1}{c}{AUC↑} & \multicolumn{1}{c}{FPR95↓} & \multicolumn{1}{c}{AUC↑} & \multicolumn{1}{c}{FPR95↓} & \multicolumn{1}{c}{AUC↑} & \multicolumn{1}{c|}{FPR95↓} & \multicolumn{1}{c}{AUC↑} & \multicolumn{1}{c}{FPR95↓} \\
		\midrule
		\multirow{10}[4]{*}{ViT} & MSP   & 91.42\% & 44.95\% & 83.92\% & 58.56\% & 73.15\% & 83.5\%  & 84.93\% & 56.63\% & 83.56\% & 58.19\% & 83.4\% & 60.37\% \\
		& Energy & 89.16\% & 56.36\% & 84.12\% & 57.23\% & 72.11\% & 89.65\% & 84.16\% & 59.95\% & 84.44\% & 56.89\% & 82.8\% & 64.02\% \\
		& Maxlogit & 91.51\% & 42.07\% & 84.61\% & 55.81\% & 72.73\% & 88.93\% & 85.41\% & 53.12\% & 84.61\% & 55.83\% & 83.77\% & 59.15\% \\
		& ODIN  & 89.74\% & 50.58\% & 90.16\% & 43.11\% & 84.15\% & 70.15\% & 87.4\% & 46.85\% & 87.15\% & 50.18\% & 87.72\% & 52.17\% \\
		& KLMatching & 96.9\%  & 10.95\% & 90.52\% & 40.32\% & 92.51\% & 58.57\% & 90.85\% & 40.85\% & 90.69\% & 60.88\% & 92.29\% & 42.31\% \\
		& Mahalanobis & 96.28\% & 10.23\% & 92.85\% & 23.38\% & 93.26\% & 25.3\% & 91.95\% & 37.46\% & 92.41\% & 38.69\% & 93.35\% & 27.01\% \\
		& ViM   & 97.99\% & 7.38\%  & 96.01\% & 14.33\% & 94.86\% & 22.91\% & 93.16\% & 35.45\% & 93.05\% & 35.91\% & 95.01\% & 23.2\% \\
		& DML   & 97.01\% & 12.42\% & 96.65\% & 14.87\% & 94.5\%  & 34.33\% & \underline{96.76\%} & \underline{16.09\%} & \underline{96.13\%} & \underline{19.6\%}  & 96.21\% & 19.46\% \\
		\cmidrule{2-14}          & C-DDS & \underline{98.69\%} & \underline{4.19\%}  & \underline{96.79\%} & \underline{10.48\%} & \underline{95.5\%}  & \underline{17.23\%} & 96.28\% & 17.85\% & 95.64\% & 20.49\% & \underline{96.58\%} & \underline{14.04\%} \\
		& C-DDS+ & \textbf{98.9\%} & \textbf{3.97\%} & \textbf{97.28\%} & \textbf{9.54\%} & \textbf{96.03\%} & \textbf{16.69\%} & \textbf{96.75\%} & \textbf{16.68\%} & \textbf{96.12\%} & \textbf{19.67\%} & \textbf{97.02\%} & \textbf{13.31\%} \\
		\bottomrule
	\end{tabular}%
	\label{ViT}%
	\vspace{1.5em}
	\centering
	\renewcommand{\arraystretch}{1.15}
	\setlength{\tabcolsep}{4.5pt}
	\caption{OOD detection of our proposed methods and the baseline methods under the Swin-B model with $Acc@1 \cite{9847099}=96.57\%$}
	\begin{tabular}{clrrrrrrrrrr|rr}
		\toprule
		\multirow{2}[2]{*}{Model} & \multicolumn{1}{c}{\multirow{2}[2]{*}{Method}} & \multicolumn{2}{c}{iNaturalist} & \multicolumn{2}{c}{Texture} & \multicolumn{2}{c}{ImageNet-O} & \multicolumn{2}{c}{Sun} & \multicolumn{2}{c|}{Places} & \multicolumn{2}{c}{average} \\
		&       & \multicolumn{1}{c}{AUC↑} & \multicolumn{1}{c}{FPR95↓} & \multicolumn{1}{c}{AUC↑} & \multicolumn{1}{c}{FPR95↓} & \multicolumn{1}{c}{AUC↑} & \multicolumn{1}{c}{FPR95↓} & \multicolumn{1}{c}{AUC↑} & \multicolumn{1}{c}{FPR95↓} & \multicolumn{1}{c}{AUC↑} & \multicolumn{1}{c|}{FPR95↓} & \multicolumn{1}{c}{AUC↑} & \multicolumn{1}{c}{FPR95↓} \\
		\midrule
		\multirow{10}[4]{*}{Swin} &MSP  & 95.67\% & 18.73\% & 88.66\% & 44.76\% & 76.15\% & 74.88\% & 89.02\% & 49.31\% & 88.78\% & 49.01\% & 87.65\% & 47.34\% \\
		& Energy & 96.51\% & 15.45\% & 92\%  & 37.23\% & 80.77\% & 68.01\% & 91.52\% & 43.42\% & 87.11\% & 51.14\% & 89.58\% & 43.05\% \\
		& Maxlogit & 96.81\% & 15.13\% & 92.96\% & 35.96\% & 83.81\% & 65.02\% & 91.95\% & 44.96\% & 89.05\% & 46.01\% & 90.91\% & 41.42\%\\
		& ODIN  & 96.48\% & 15.79\% & 92.92\% & 35.41\% & 87.38\% & 54.27\% & 90.07\% & 47.04\% & 88.93\% & 48.63\% & 91.16\% & 40.23\% \\
		& KLMatching & 97.91\% & 12.33\% & 91.47\% & 38.45\% & 92.86\% & 48.37\% & 94.34\% & 34.3\% & 93.23\% & 38.52\% & 93.96\% & 34.39\%\\
		& Mahalanobis & \textbf{99.53\%} & \textbf{2.19\%} & 96.21\% & 11.88\% & 95.07\% & 29.98\% & 96.85\% & 14.83\% & 96.19\% & 17.99\% & 96.77\% & 15.37\% \\
		& ViM   & 98.46\% & 6.1\% & \underline{97.8\%} & 8.53\%  & 95.45\% & 26.41\% & \underline{97.16\%} & 14.44\% & 96.68\% & 16.31\% & 97.11\% & 14.36\% \\
		& DML   & 93.58\% & 48.59\% & 94.99\% & 31.62\% & 94.36\% & 44.39\% & 95.11\% & 32.35\% & 96.12\% & 20.98\% & 94.83\% & 35.59\% \\
		\cmidrule{2-14}          & C-DDS & 98.98\% & \underline{3.64\%}  & 97.67\% & \underline{6.46\%} & \underline{96.42\%} & \underline{13.87\%} & 96.73\% & \underline{13.82\%} & \underline{97.49\%} & \underline{10.26\%} & \underline{97.46\%} & \underline{9.61\%}\\
		& C-DDS+  & \underline{99.04\%} & 3.72\%  & \textbf{98.32\%} & \textbf{5.94\%} & \textbf{97.23\%} & \textbf{13.4\%} & \textbf{97.5\%} & \textbf{10.53\%} & \textbf{98.26\%} & \textbf{7.72\%} & \textbf{98.07\%} & \textbf{8.26\%} \\
		\bottomrule
	\end{tabular}%
	\label{Swin}%
\end{table*}%

\subsection{Impact of common features on C-DDS}
\begin{figure}[!h]
	\centering
	\includegraphics[width=3in]{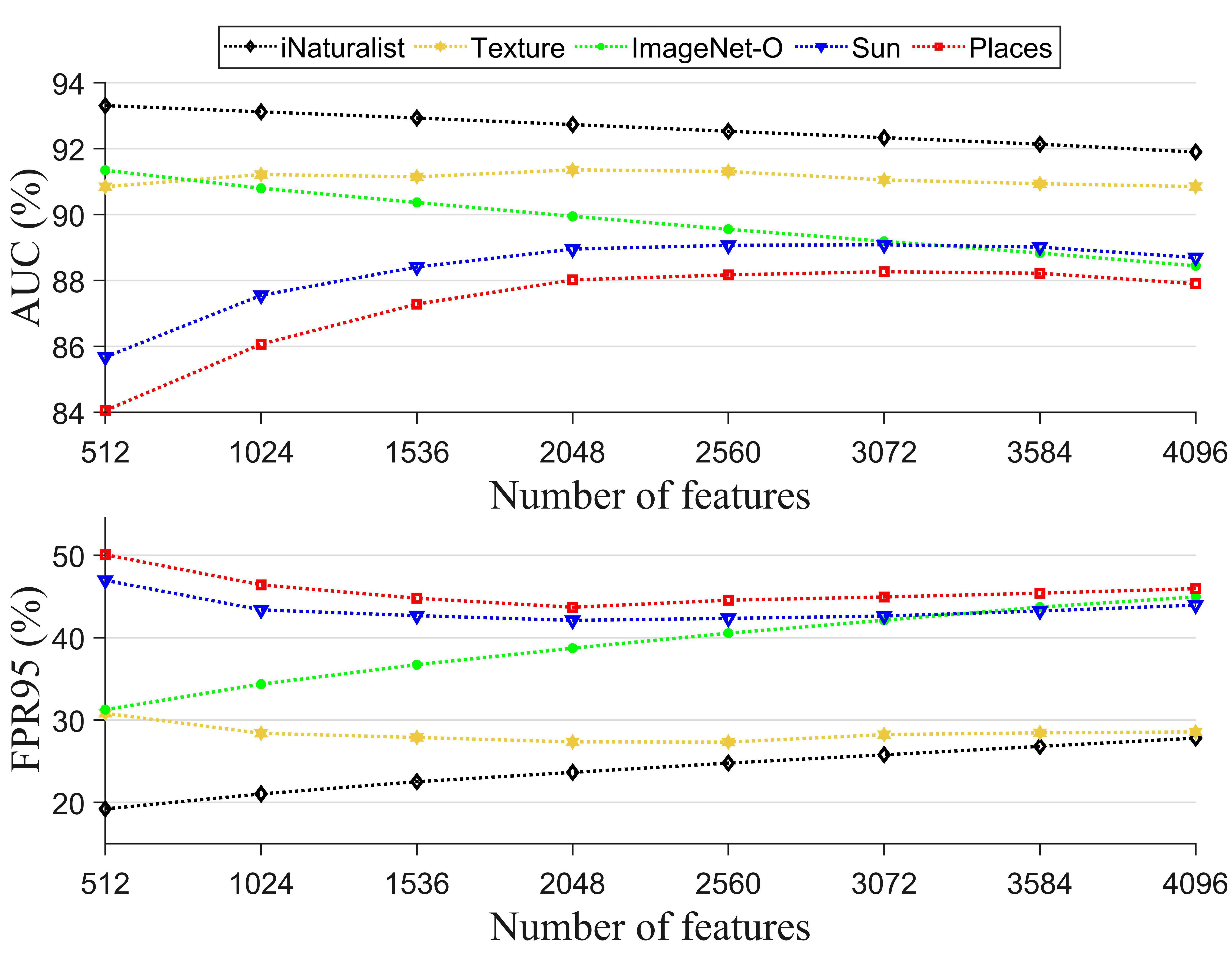}%
	\caption{The impact of varying numbers of common features on the performance of C-DDS.}
	\label{features}
\end{figure}

The extraction of feature vectors significantly influences the performance of C-DDS. An abundance of shared features can potentially introduce biases and misguide the decision-making process of C-DDS methodologies. To address this, we investigate the number of common features among feature vectors obtained through the utilization of the VGG model. The outcomes of our analysis are presented in Fig \ref{features}. Our findings indicate that as the number of common features increases, the AUC exhibits a declining trend for both the ImageNet-O and iNaturalist datasets. This suggests that an excessive amount of shared features indeed impacts the decision-making capabilities of C-DDS. Conversely, when examining the Sun and Place datasets, we observe that the AUC increases with the number of common features until a plateau is reached. This observation implies that an insufficient number of shared features can also be detrimental to the decision-making process of C-DDS. Notably, in the Texture dataset, we observe a rising trend in AUC until 2560 common features are retained, followed by a subsequent decline. This pattern suggests that an optimal number of common features exists for C-DDS performance in the Texture domain, which can get the best performace of C-DDS.

\subsection{Impact of scaling hyper-parameters on C-DDS+}
\begin{figure*}[!h]
	\centering
	\subfloat[]{\includegraphics[width=3in]{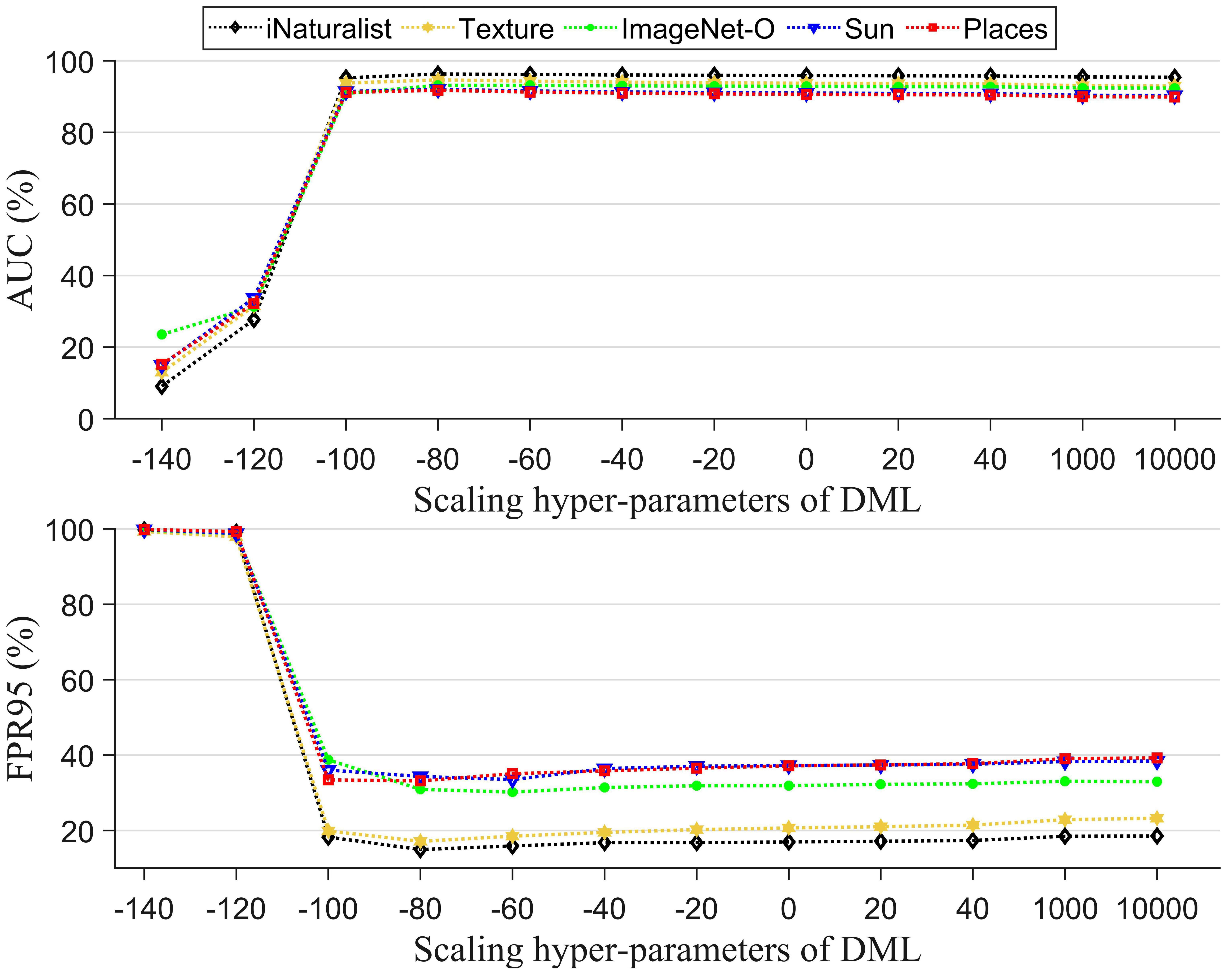}%
	}
	\hfil
	\subfloat[Some scaling hyper-parameters in (a)]{\includegraphics[width=3.12in]{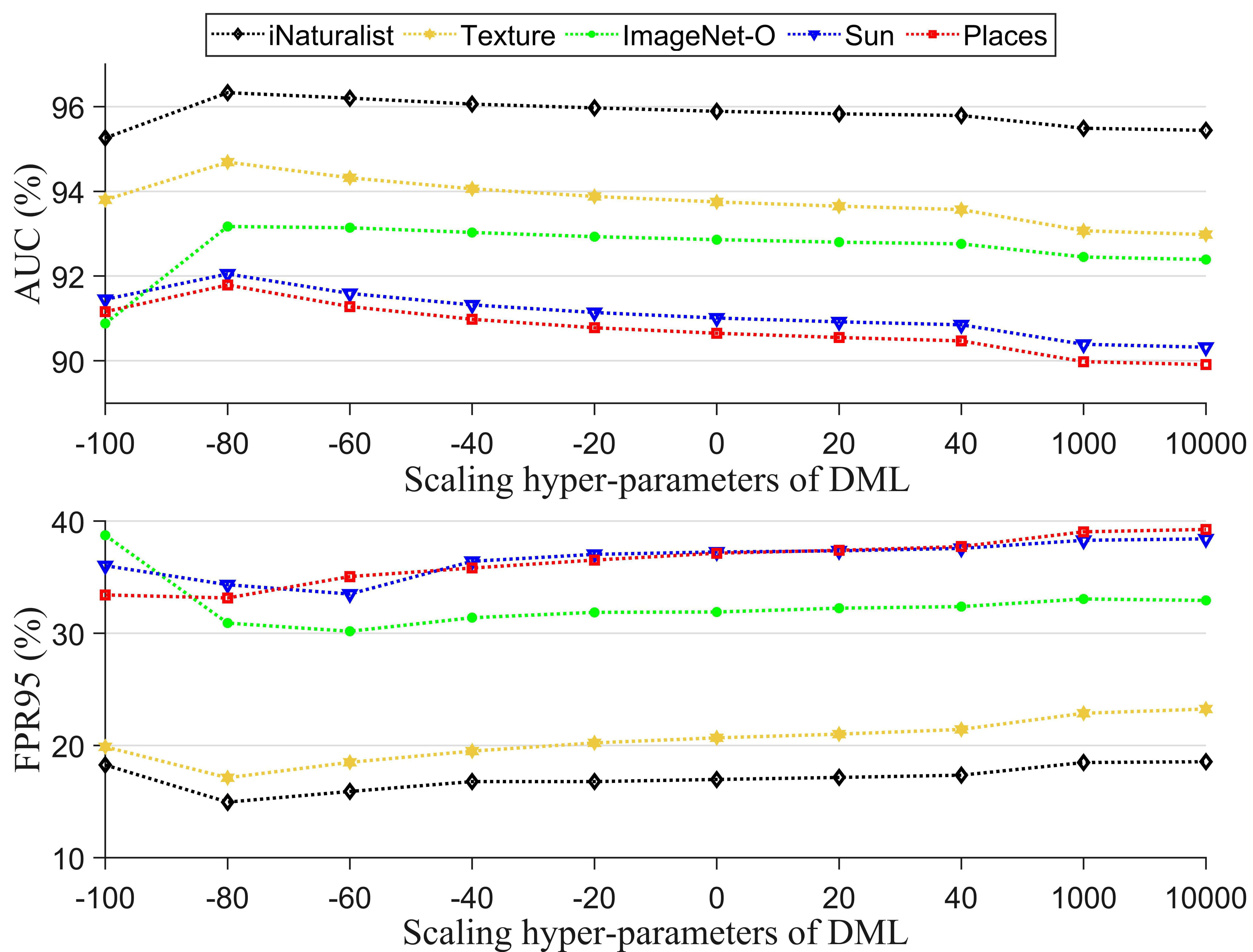}%
	}
	\caption{The impact of scaling hyper-parameter on the performance of C-DDS+.}
	\label{hyper}
\end{figure*}

The scaling hyper-parameter plays a vital role in balancing the scores of C-DDS and DML in our approach. Given that the OOD data situation is often unknown, we recommend assigning equal importance to both components. Fine-tuning these scaling hyper-parameters can effectively adapt to specific types of OOD datasets. Notably, as the scaling hyper-parameter $k_1$ remains consistently set at $0$, we focus on determining the impact of different scaling hyper-parameters $k_2$ on C-DDS+ under the VGG model. A higher value of $k_2$ emphasizes the significance of logarithmic information, whereas a lower value assigns greater weight to features and softmax probabilities. Fig. \ref{hyper} revealed that setting $k_2=-82$ yielded better results. Furthermore, a linear reduction of the variable $k_2$ introduces the inclusion of negative scores within the DML, thereby inducing a huge decline in the AUC. This reduction, although having a minimal impact on DML's decision-making capabilities, causes DML to deviate from the definition of mass function, rendering it incompatible with the PCR5. Conversely, increasing the value of $k_2$ gradually diminished the AUC of C-DDS+ and eventually, the AUC converged to that of C-DDS, primarily because the scores in DML approached unity as $k_2$ increased, thereby weakening the influence of DML on decision making. Notably, each dataset exhibited an optimal pair of parameters that maximized the performance of C-DDS+.

\section{Conclusion}
This paper investigates a novel quantum conflict measure and its application in out-of-class identification. Initially, we introduce the QCC and the QCI derived from QCC, aiming to facilitate decision-making in uncertain environments. Additionally, we define and analyze the properties of QCC and QCI, confirming that QCI possesses desirable attributes such as non-negativity, symmetry, boundedness, extreme consistency, and insensitivity to refinement. Subsequently, we integrate QCI into conflict fusion and compare its performance with several conventional fusion methods using the UCI dataset. The results demonstrate certain advantages of QCI over these fusion methods. Furthermore, to explore the potential of QCI, we apply a QCI-based fusion approach to the task of OOD detection. We propose the C-DDS method and its updated version, C-DDS+, which leverage both QCI and our earlier methodologies. Remarkably, C-DDS+ exhibits significant superiority over several state-of-the-art OOD detection methods on the Imagenet dataset. These findings further affirm the excellent utility and wide-ranging application potential of QCI. In future research, we aim to enhance the efficiency of QCI by integrating it with quantum parallel computing techniques and explore further applications of this measure.

\begin{footnotesize}

\end{footnotesize}

\clearpage
\begin{small}
\appendices
\section{The relevant proofs of the properties of QCC}\label{QCC_p}
\subsection{Proof of the definition of the inner product in Eq (\ref{innerproduct}):}
To prove that Eq (\ref{innerproduct}) is an inner product, i.e., to fulfill the following three conditions:
\begin{enumerate}
	\item{\textit{Conjugate symmetry:} $\left\langle qm_1\middle|qm_2\right\rangle=\overline{\left\langle qm_1\middle|qm_2\right\rangle}$.}
	\item{\textit{Conjugate Linear:} $z\left\langle qm_1\middle|qm_2\right\rangle=\overline{z}\left\langle qm_1\middle|qm_2\right\rangle$ and $\left\langle qm_1 +qm_3\middle|qm_2\right\rangle=\left\langle qm_1\middle|qm_2\right\rangle+\left\langle qm_3\middle|qm_2\right\rangle$.}
	\item{\textit{Positivity:} $\left\langle qm_1\middle|qm_1\right\rangle=0$, iff $qm_1$ is a zero vector.}
\end{enumerate}
Here, $qm_1$, $qm_2$ and $qm_3$ represent three QMFs on $2^{\left|\Theta\right\rangle}=\left\lbrace\emptyset,A_1,... ,A_{2^n-1}\right\rbrace$ and $z$ is an arbitrary complex number.\\
\textbf{Proof} (Conjugate symmetry): 
\begin{align}
	\left\langle qm_1\middle|qm_2\right\rangle=\sum_{i=1}^{2^n}\sum_{j=1}^{2^n}{\overline{qm_1(A_i)}qm_2(A_j)\mathcal{G}(\dot{\theta}_{i,j})\frac{|A_i\cap A_j|}{|A_i\cup A_j|}},\notag\\
	\overline{\left\langle qm_2\middle|qm_1\right\rangle}=\sum_{i=1}^{2^n}\sum_{j=1}^{2^n}{qm_2(A_i)\overline{qm_1(A_j)}\mathcal{G}(\dot{\theta}_{i,j})\frac{|A_i\cap A_j|}{|A_i\cup A_j|}}.\notag
\end{align}
It is clear to see that $\left\langle qm_1\middle|qm_2\right\rangle=\overline{\left\langle qm_2\middle|qm_1\right\rangle}$.\\
\textbf{End of proof}\\
\textbf{Proof} (Conjugate Linear):
\begin{align}
	&\left\langle{zqm}_x\middle|qm_2\right\rangle=\sum_{i=1}^{2^n}\sum_{j=1}^{2^n}{\overline{{{zqm}_x(\mathcal{F}}_i)}{qm_2(\mathcal{F}}_j)\mathcal{G}(\dot{\theta}_{i,j})\frac{\left|A_i\cap A_j\right|}{\left|A_i\cup A_j\right|}}=\notag\\
	&\bar{z}\sum_{i=1}^{2^n}\sum_{j=1}^{2^n}{\overline{{qm_1(\mathcal{F}}_i)}\ {qm_2(\mathcal{F}}_j)\mathcal{G}(\dot{\theta}_{i,j})\frac{\left|A_i\cap A_j\right|}{\left|A_i\cup A_j\right|}}=\bar{z}\left\langle qm_1\middle|qm_2\right\rangle.\notag\\
	&\notag\\
	&\left\langle qm_1 +qm_3\middle|qm_2\right\rangle\notag\\
	&=\sum_{i=1}^{2^n}\sum_{j=1}^{2^n}{[\overline{qm_1(A_i)+qm_3(A_i)}] {qm_2(\mathcal{F}}_j)\mathcal{G}(\dot{\theta}_{i,j})\frac{|A_i\cap A_j|}{|A_i\cup A_j|}}\notag\\
	&=\sum_{i=1}^{2^n}\sum_{j=1}^{2^n}{[\overline{qm_1(A_i)}+\overline{qm_3(A_i)}] {qm_2(\mathcal{F}}_j)\mathcal{G}(\dot{\theta}_{i,j})\frac{|A_i\cap A_j|}{|A_i\cup A_j|}}\notag\\
	&=\left\langle qm_1\middle|qm_2\right\rangle+\left\langle qm_3\middle|qm_2\right\rangle.\notag
\end{align}
\textbf{End of proof}\\
\textbf{Proof} (Positivity):
\begin{align}
	\left\langle qm_1\middle|qm_1\right\rangle=\sum_{i=1}^{2^n}\sum_{j=1}^{2^n}{\overline{{qm_1(\mathcal{F}}_i)}{qm_1(\mathcal{F}}_j)\mathcal{G}(\dot{\theta}_{i,j})\frac{|A_i\cap A_j|}{|A_i\cup A_j|}}.\notag
\end{align}
Given that we have defined the range of the angular phase as $[0, \frac{\pi}{2}]$, it becomes evident that $\mathcal{G}(\dot{\theta}_{i,j})\geq0$ and $\left\langle qm_1\middle|qm_1\right\rangle\geq0$. To strengthen our argument, it is advantageous to assign a value of $z$ to $qm_1(A_i)$, where $z\neq0$, while keeping the remaining elements as zero. Consequently, $\left\langle qm_1\middle|qm_1\right\rangle=\bar{z}z$. Iff $z=0$, $\left\langle qm_1\middle|qm_1\right\rangle=0$. This finding contradicts our initial conclusion, leading us to the deduction that $\left\langle qm_1\middle|qm_1\right\rangle=0$, iff $qm_1$ is a zero vector.\\
\textbf{End of proof}\\
\subsection{Proof of the properties of QCC:}
\noindent\textbf{Proof} (Non-negativity): For two arbitrary QMFs $qm_1$ and $qm_2$ on $2^{\left|\Theta\right\rangle}=\left\lbrace\emptyset,A_1,... ,A_{2^n-1}\right\rbrace$, we have
\begin{align}
	QCC(qm_1,qm_2)=\left[\frac{|\left\langle qm_1\middle|qm_2\right\rangle|}{||qm_1||\ ||qm_2||}\right]^4.\notag
\end{align}
Clearly, $QCC(qm_1,qm_2)$ cannot be negative.\\
\textbf{End of proof}\\
\noindent\textbf{Proof} (Symmetry): For two arbitrary QMFs $qm_1$ and $qm_2$ on $2^{\left|\Theta\right\rangle}=\left\lbrace\emptyset,A_1,... ,A_{2^n-1}\right\rbrace$, we have
\begin{align}
	\left[\frac{|\left\langle qm_1\middle|qm_2\right\rangle|}{||qm_1||\cdot||qm_2||}\right]^4=\left[\frac{|\overline{\left\langle qm_2\middle|qm_1\right\rangle}|}{||qm_2||\cdot||qm_1||}\right]^4=\left[\frac{|\left\langle qm_2\middle|qm_1\right\rangle|}{||qm_2||\cdot||qm_1||}\right]^4\notag
\end{align}
\textbf{End of proof}\\
\noindent\textbf{Proof} (Boundedness): For this property, we need to prove the following two propositions separately:

(1) $QCC(qm_1,qm_2)\geq 0$

(2) $QCC(qm_1,qm_2)\leq 1$\\
\textit{Proof of} (1): Clearly, $QCC(qm_1,qm_2)\geq 0$ can be proved due to the non-negativity of $QCC$.\\
\textit{End proof of} (1)\\
\textit{Proof of} (2): for two arbitrary QMFs $qm_1$ and $qm_2$ on $2^{\left|\Theta\right\rangle}=\left\lbrace\emptyset,A_1,... ,A_{2^n-1}\right\rbrace$, Eq (\ref{QCC}) can be rewritten in the following form:
\begin{align}
	QCC(qm_1,qm_2)=\frac{|qm_1^\dagger G^\top G qm_2|^4}{|qm_1^\dagger G^\top G qm_1|^2\cdot|qm_2^\dagger G^\top G qm_2|^2}\notag,
\end{align}
where ${qm_1=\left[qm_1(\emptyset),...,qm_1(A_{2^n-1})\right]^\top}$ and $qm_1^\dagger$ represents the conjugate transpose of $qm_1$, idem $qm_2$. Here, the matrix $G$ represents a lower triangular matrix \cite{xiao2020novel} and $G^TG$ denotes a symmetric positive definite matrix related to $\mathcal{G}(\dot{\theta}_{i,j})$ and $\frac{|A_i\cap A_j|}{|A_i\cup A_j|}$. Therefore, from the triangle inequality, we can derive:
\begin{align}
	|G(qm_1+qm_2)|^2\leq(|Gqm_1|+|Gqm_2|)^2.\label{triangle}
\end{align}
Since there exists $|qm_1^\dagger G^\top G qm_2|=|qm_2^\dagger G^\top G qm_1|$ and $qm_1^\dagger G^\top G qm_1=(G\overline{qm_1})^\top\cdot Gqm_1=|Gqm_1|^2$, we have
\begin{align}
	\Longleftrightarrow&|(qm_1+qm_2)^\dagger G^\top G (qm_1+qm_2)|\leq|qm_1^\dagger G^\top G qm_1|+\notag\\
	&|qm_2^\dagger G^\top G qm_2|+2\sqrt{|qm_1^\dagger G^\top G qm_1|\cdot|qm_2^\dagger G^\top G qm_2|}\notag\\
	\Longleftrightarrow&|qm_1^\dagger G^\top G qm_1|+|qm_2^\dagger G^\top G qm_2|+|qm_2^\dagger G^\top G qm_1|+\notag\\
	&|qm_1^\dagger G^\top G qm_2|\leq|qm_1^\dagger G^\top G qm_1|+|qm_2^\dagger G^\top G qm_2|+\notag\\
	&2\sqrt{|qm_1^\dagger G^\top G qm_1|\cdot|qm_2^\dagger G^\top G qm_2|}\notag\\
	\Longleftrightarrow&\frac{|qm_1^\dagger G^\top G qm_2|^2}{|qm_1^\dagger G^\top G qm_1|\cdot|qm_2^\dagger G^\top G qm_2|}\leq1\notag
\end{align}
Clearly, $QCC(qm_1,qm_2)\leq 1$ can be conducted.\\
\textit{End proof of} (2)\\
\textbf{End of proof}\\
\textbf{Proof} (Nondegeneracy): Based on Eq (\ref{triangle}), we can establish the following theorem: The condition for the equality sign in the triangular inequality is exclusively satisfied when the three points are collinear. It is important to note that $qm_1$ and $qm_2$ are collinear only when $qm_1$ is equal to $qm_2$. As a result, the presence of $QCC(qm_1,qm_2)=1$ can only occur under the condition that $qm_1=qm_2$.\\
\textbf{End of proof}

\section{The relevant proofs of the properties of QCI}\label{QCI_p}
Having established the properties of nondegeneracy, symmetry, and boundedness for QCC, it becomes evident that these properties are also satisfied for QCI. Furthermore, since the definition of QCI in this paper does not consider any aspects related to $2^{\left|\Theta\right\rangle}$, it follows that the proof of Insensitivity to refinement is of trivial nature. Hence, the primary attention is directed towards investigating the property of Extreme consistency for QCI.\\
\noindent\textbf{Proof} (Extreme consistency): To address this property adequately, it is necessary to provide separate proofs for the following two propositions:

(1) $QCI(qm_1,qm_2)=1$, iff for $A_i$ and $A_j$ of $qm_1$ and $qm_2$ respectively, $(\cup A_i)\cap(\cup A_j)=\emptyset$,

(2) $QCI(qm_1,qm_2)=0$, iff $qm_1=qm_2$.\\
\textit{Proof of} (1): For two arbitrary QMFs $qm_1$ and $qm_2$ on $2^{\left|\Theta\right\rangle}=\left\lbrace\emptyset,A_1,... ,A_{2^n-1}\right\rbrace$, let us suppose that there exist non-zero elements $qm_1(A_1)\neq 0$ and $qm_2(A_1)\neq 0$ in $qm_1$ and $qm_2$, respectively. Additionally, it is assumed that there is no intersection between any other elements. Therefore, we have:
\begin{align}
	QCI(qm_1,qm_2)=1-\left[\frac{|\left\langle qm_1\middle|qm_2\right\rangle|}{||qm_1||\cdot||qm_2||}\right]^4.\notag
\end{align}
Since both $||qm_1||$ and $||qm_2||$ are greater than zero and we have
\begin{align}
	\left\langle qm_1\middle|qm_2\right\rangle=\overline{qm_1(A_1)}qm_2(A_1)\mathcal{G}(\dot{\theta_{1,1}}),\notag
\end{align}
and
\begin{align}
	QCI(qm_1,qm_2)=1-\left[\frac{|\overline{qm_1(A_1)}qm_2(A_1)\mathcal{G}(\dot{\theta_{1,1}})|}{||qm_1||\cdot||qm_2||}\right]^4 < 1.\notag
\end{align}
Therefore, $QCI(qm_1,qm_2)=1$, iff for $A_i$ and $A_j$ of $qm_1$ and $qm_2$ respectively, $(\cup A_i)\cap(\cup A_j)=\emptyset$.\\
\textit{End proof of} (1)\\
\textit{Proof of} (2): By utilizing the Nondegeneracy of QCC, we can deduce that proposition $QCI(qm_1,qm_2)=0$ is valid iff $qm_1=qm_2$.\\
\textit{End proof of} (2)\\
\textbf{End of proof}
\end{small}
\vfill
\end{CJK}

\end{document}